\documentclass[10pt,twocolumn,letterpaper]{article}

\usepackage{iccv}
\usepackage{times}
\usepackage{epsfig}
\usepackage{graphicx}
\usepackage{amsmath}
\usepackage{amssymb}
\usepackage[accsupp]{axessibility}  

\usepackage{color}
\usepackage{url}
\usepackage{bm}
\usepackage{mathtools}
\usepackage{amsfonts} 
\usepackage{amsthm}
\usepackage{array,multirow}
\usepackage{tabularx}
\usepackage{arydshln}
\usepackage{adjustbox}
\usepackage{wrapfig}
\usepackage{xcolor, pifont}
\usepackage{soul}
\usepackage{booktabs}

\newcommand{\cmark}{\ding{51}}%
\newcommand*\colourcheck[1]{%
  \expandafter\newcommand\csname #1check\endcsname{\textcolor{#1}{\ding{51}}}%
}
\newcommand*\colourx[1]{%
  \expandafter\newcommand\csname #1x\endcsname{\textcolor{#1}{\ding{55}}}%
}
\colourcheck{blue}
\colourcheck{green}
\colourcheck{gray}
\colourx{gray}

\theoremstyle{plain}
\newtheorem{theorem}{Theorem}[section]

\newtheorem{remark}[theorem]{Remark}
\theoremstyle{remark}

\newcommand{\hjs}[1]{\textcolor{black}{#1}}
\newcommand{\jh}[1]{\textcolor{black}{#1}}

\usepackage[pagebackref=true,breaklinks=true,letterpaper=true,colorlinks,bookmarks=false]{hyperref}

\iccvfinalcopy 

\def\iccvPaperID{12573} 

\ificcvfinal\fi

\begin{document}

\title{PC-Adapter: Topology-Aware Adapter for Efficient Domain Adaption \\ on Point Clouds  with Rectified Pseudo-label}

\author{Joonhyung Park \\
KAIST \\
{\tt\small deepjoon@kaist.ac.kr}
\and Hyunjin Seo \\
KAIST \\
{\tt\small bella72@kaist.ac.kr}
\and Eunho Yang \\
KAIST, AITRICS \\
{\tt\small eunhoy@kaist.ac.kr}
}

\maketitle
\ificcvfinal\thispagestyle{empty}\fi

\begin{abstract}
Understanding point clouds captured from the real-world is challenging due to shifts in data distribution caused by varying object scales, sensor angles, and self-occlusion. Prior works have addressed this issue by combining recent learning principles such as self-supervised learning, self-training, and adversarial training, which leads to significant computational overhead. 
Toward succinct yet powerful domain adaptation for point clouds, we revisit the unique challenges of point cloud data under domain shift scenarios and discover the importance of the global geometry of source data and trends of target pseudo-labels biased to the source label distribution. Motivated by our observations, we propose an adapter-guided domain adaptation method, \textbf{PC-Adapter}, that preserves the global shape information of the source domain using an attention-based adapter, while learning the local characteristics of the target domain via another adapter equipped with graph convolution. Additionally, we propose a novel pseudo-labeling strategy resilient to the classifier bias by adjusting confidence scores using their class-wise confidence distributions to consider relative confidences. Our method demonstrates superiority over baselines on various domain shift settings in benchmark datasets - PointDA, GraspNetPC, and PointSegDA.

\end{abstract}

\section{Introduction}
3D vision has gained considerable attention as an immense amount of 3D data is collected by LiDAR sensors or depth-sensing cameras in real-world applications such as autonomous vehicles, surveillance systems, and drones. In recent years, deep neural networks~\cite{pointnet,pointnet++,kpconv,pointtransformer,pointnext} have become a de facto approach for understanding unordered 3D point sets (\ie point cloud data). However, for the case of point clouds in the wild, deep models inevitably encounter \textit{distributional shift} - also known as \textit{domain shift} - due to the different object scales, point density, angle of sensor, object occlusion, \textit{etc}. 
This problem makes the models perform poorly outside the laboratory.

Unsupervised domain adaptation (UDA) is a representative solution to tackle the domain shift problem, which targets to transfer the knowledge from a labeled source domain to an unlabeled target domain~\cite{adda,cycada,kang2019contrastive}. UDA methods generally endeavor to reduce the discrepancy of two domains in representation space, so that a classifier trained on the label-rich source domain generalizes well on the target domain. Such standard UDA strategies would be attractive options even for point cloud data, since collecting synthetic point clouds can be simply done by sampling points from the 3D CAD models whereas annotating the enormous real 3D data is cumbersome and time-consuming. Hence, if a UDA method can successfully transfer the knowledge from synthetic point clouds (source domain) to real point clouds (target domain), it will enable the models to significantly boost performances over a broad range of 3D vision tasks.

Therefore, numerous studies considering the innate characteristic of point cloud data have recently been proposed for domain adaptation~\cite{pointdan, defrec, gast, glrv, shen2022domain, mlsp, rs}. Most of them \textit{jointly} exploit components such as self-supervised learning, adversarial training, data augmentation, and self-training to squeeze the best adaptation performances. Especially, recent works~\cite{gast,glrv,mlsp} focus on designing at least two self-supervised learning tasks by deforming point clouds; for example, reconstructing the deformed parts or predicting the extent of specific distortion.
Although these methods have shown successful results, due to their excessive computations and memory access, these frameworks are difficult to deploy in real-world systems where we do not have sufficient computing resources. 
These situations raise questions regarding the race of performance-oriented design and call for a pivot towards a succinct yet effective strategy toward a practical domain adaptation framework. 

Toward this, we reconsider the individual efficacy of different types of knowledge for domain adaptation, especially focusing on which knowledge of the source benefits to be transferred, and suggest two design philosophies for a point cloud-specific domain adaptation framework. 
We first compare the value of knowledge in terms of global geometry and local structure to determine which knowledge would be more beneficial under domain shift scenarios. Drawing on our observations, we argue the necessity of transferring the shape information of the source domain during the adaptation phase. 
Secondly, we observe that pseudo-labels 
are highly dependent on the label distribution of the source. To obtain unbiased pseudo-labels for target data, we suggest that the label bias induced from the knowledge of source label distribution should be considered in selecting pseudo-labels as there is no guarantee that the label distribution of the source domain would match that of the target domain.


Inspired by the prior insights, we introduce \textbf{P}oint \textbf{C}loud Adapter (\textit{PC-Adapter}), an adapter-based domain adaptation framework that can efficiently learn feature transformation to adapt local spatial structures of the target domain while preserving the knowledge of global geometry acquired from the source domain. Our method exploits two adapters, a global shape-aware adapter and a locality-aware adapter, both of which consist of one simple block. Our global shape-aware adapter learns implicit shape information of the source domain using a proposed relative positional encoding and then transfers this knowledge to target point clouds. To preserve the source geometry information during the adaptation process, the parameters are weakly updated when training target point clouds. Meanwhile, the locality-aware adapter actively learns the target domain-specific local structures using graph neural network operations, which is \textit{only} updated by target point clouds. In other words, the local structure adapter solely provides room for feature transformation to adapt to the target domain. After passing the entire path of the model including adapters, target point clouds are trained via pseudo-labels. To avoid pseudo-labels being selected relying on the label distribution of the source domain, we propose a simple yet novel pseudo-label correction strategy. Our approach rectifies pseudo-labels based on their percentile ranks from approximated class-wise confidence distributions, taking into account their relative confidences. 


In summary, our contribution is threefold:
\begin{itemize}
    \item We present design philosophies to promote succinct yet effective domain adaptation on point cloud data, concentrated on which knowledge from the source domain has to be transferred. 
    \item We propose a tailored framework for these design principles that
   preserves global shape knowledge of the source via an attention-based adapter while providing room for target-specific feature transformation through a graph convolution-based adapter. We also devise a reliable pseudo-labeling method that adjusts class-wise confidence with the guidance of its percentile on confidence distribution. Note that our method does not resort to any extra sub-tasks or adversarial training. 
    \item Extensive experiments show that our \textit{PC-Adapter} brings significant performance gain over self-training and adversarial domain alignment baselines,  
    and even achieves state-of-the-art performances when combined with auxiliary self-supervised learning tasks. 
\end{itemize}

\section{Related Work}
\subsection{Domain Adaptation on Point Clouds}
UDA has been recently investigated on 3D point clouds to bridge the gap between domains of different natures.
PointDAN~\cite{pointdan} proposed a point cloud-specialized UDA framework via multi-scale feature alignment.
DefRec~\cite{defrec} improved the performance by leveraging a self-supervised learning task of reconstructing deformed object regions along with a mixup strategy.
GAST~\cite{gast} learned a domain-shared point cloud representation by integrating geometry-aware auxiliary tasks of angle/location prediction, and further encouraged the training through the pseudo-labeling process.
GLRV~\cite{glrv} devised different self-supervised learning methods to learn global and local point cloud structures in-depth combined with adversarial training and voting-based pseudo-labeling strategy.
Shen et al.~\cite{shen2022domain} learned unsigned distance fields of point clouds in a self-supervised manner to enhance the understanding of implicit shape geometry shared across domains.
MLSP~\cite{mlsp} incorporated prediction tasks of estimating local attributes of masked regions to alleviate domain bias for target point clouds.
However, most of these works generally leverage cumbersome tasks that are infeasible in real-world systems.
Different from existing studies, we propose an efficient domain adaptation framework for point clouds without requiring additional tasks with excessive computation.

\subsection{Adapter}
Recently, adapter modules have been proposed as a parameter-efficient method to fine-tune large-scale pretrained models for downstream tasks~\cite{adapter,unified_adapter,vl_adapter, vision_adapter}. The adapter modules are inserted between layers of a pretrained network, which consist of only a few trainable parameters (\eg two linear layers with nonlinear activation). This adapter-based transfer has shown successful performances on various downstream tasks by enabling the adapter module to learn the task-specific feature transformations. Inspired by the success of adapter-tuning, we exploit the concept of adapter to address domain shift problems in point cloud data. Compared to an adapter in the transfer learning field, our proposed adapter is specially designed to learn the structural information of 3D objects.
\begin{figure}[t]
\vspace{-0.00in}
  \centering
  \includegraphics[width=0.45\textwidth]{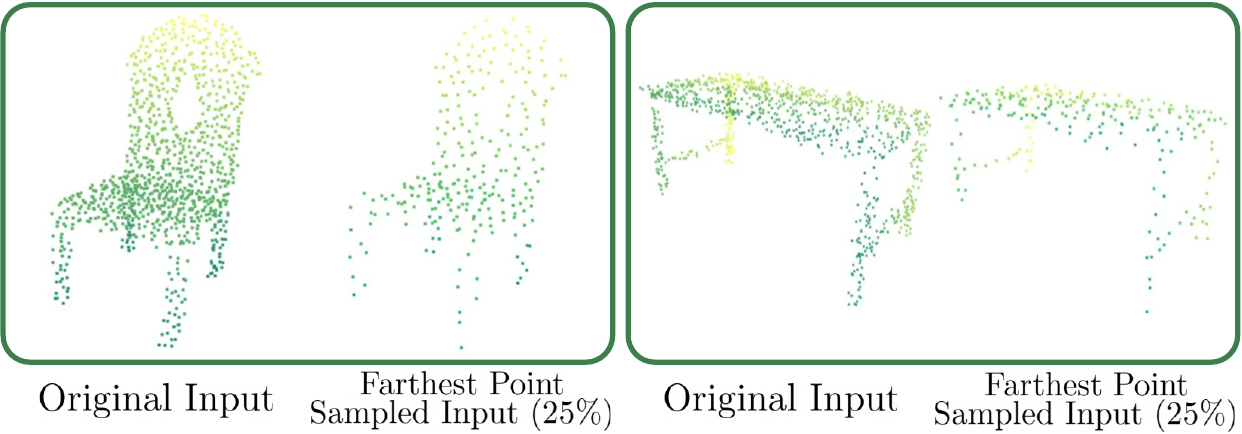}
  \includegraphics[width=0.45\textwidth]{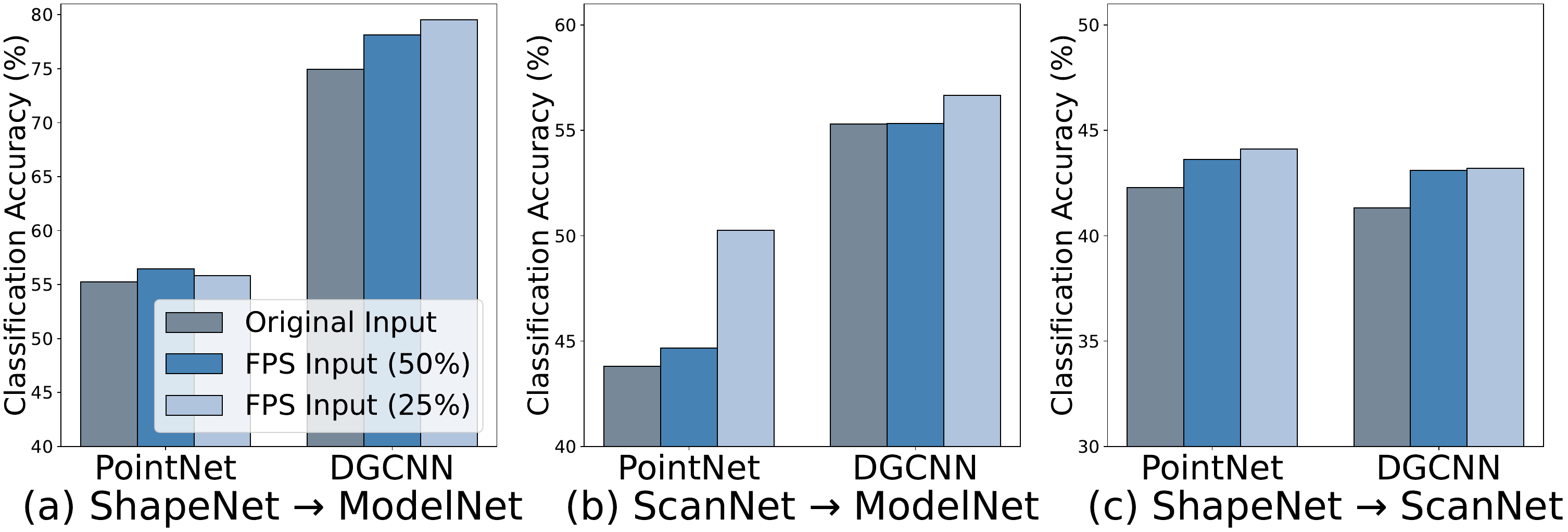}
  \caption{\small{Classification accuracy of varying input point ratios on two point cloud encoding architectures. We gradually reduce the number of given point cloud points by $50\%$ and $25\%$ using FPS, preserving only the global anatomy of inputs. Domain adaptations solely based on global structures show superior performance against the original input in multiple domain shift scenarios.}
  \vspace{-0.02in}
  }

  \label{fig:global_local_structure}
\end{figure}
\section{Design Philosophies for Domain Adaptation on Point Cloud}~\label{sec:prelim_study}
Before devising the domain adaptation method, we would like to suggest key philosophies that need to be considered for domain adaptation on point clouds.
Especially, we mainly focus on which knowledge from the source domain has to be \textit{preserved} and which information should be \textit{transformed} adaptively towards the target domain.


\vspace{0.06in}
\subsection{Problem Formulation and Notation}
In this work, we tackle unsupervised domain adaptation (UDA) for 3D point cloud data. Let $\bm{X}_{k}\in \mathbb{R}^{m\times3}$ be the point cloud data represented as an unordered set of 3D coordinates $\bm{X}_{k}= \{\mathbf{p}_1,\dots \mathbf{p}_m\}$, where $m$ is the number of points and $\mathbf{p}_i\in\mathbb{R}^3$ is a coordinate of $i$-th point. The objective of UDA is to transfer the knowledge from the annotated source domain $\mathcal{S}=\{(\bm{X}_{k}^{\text{src}},y_{k}^{\text{src}})\}_{k=1}^{n_s}$ to unlabeled target domain $\mathcal{T}=\{\bm{X}_{k}^{\text{trgt}}\}_{k=1}^{n_t}$ where $n_s$ and $n_t$ denote the number of point clouds in the source and target domains, respectively. In UDA settings, two domains are assumed to have different data distributions $\mathcal{D}_{\text{source}}$ and $\mathcal{D}_{\text{target}}$ but with a shared label space $\mathcal{Y} = \{1, \ldots, c \}$. The core concept of the UDA is to learn a model that projects inputs from different domains onto a shared representation space by reducing the discrepancy of representations across domains.

\vspace{0.06in}
\subsection{Efficacy of Global Geometry Knowledge}\label{subsec:remark3.1}
We begin our discussion by showing the importance of leveraging global geometry knowledge where local spatial structures are usually discarded in domain-shift scenarios. Given the causes of distributional shifts (\eg self-occlusion, sensor noise, point density), there generally exists a large characteristic discrepancy appearing locally across domains. Therefore, it is important to \textit{selectively} transfer the pertinent knowledge to the distribution-shifted domain rather than ignorantly transferring all types of knowledge from the source domain. To identify this `versatile' knowledge in domain shift scenarios, we compare the efficacy of global geometry information with that of local spatial information on domain-shifted point clouds. 

Toward this, we design an experiment to train models with data that only contains information about global architecture by eliminating the local attributes of point clouds using Farthest Point Sampling (FPS). Two representative models - PointNet~\cite{pointnet} and DGCNN~\cite{dgcnn} - are trained with farthest point sampled data from labeled source datasets and tested on the unlabeled target dataset without applying any domain adaptation methods. 
In Figure~\ref{fig:global_local_structure}, we can somehow surprisingly confirm that results with eliminating some points (lightening the knowledge by eliminating local properties) show superior, at least comparable, object classification accuracy compared to the original input under multiple domain shifts. These results imply that the knowledge of global geometry remains helpful when the domain shifts, whereas the local characteristics of objects do not. 
\begin{figure}[t] 
\centering
   \includegraphics[width=0.48\textwidth]{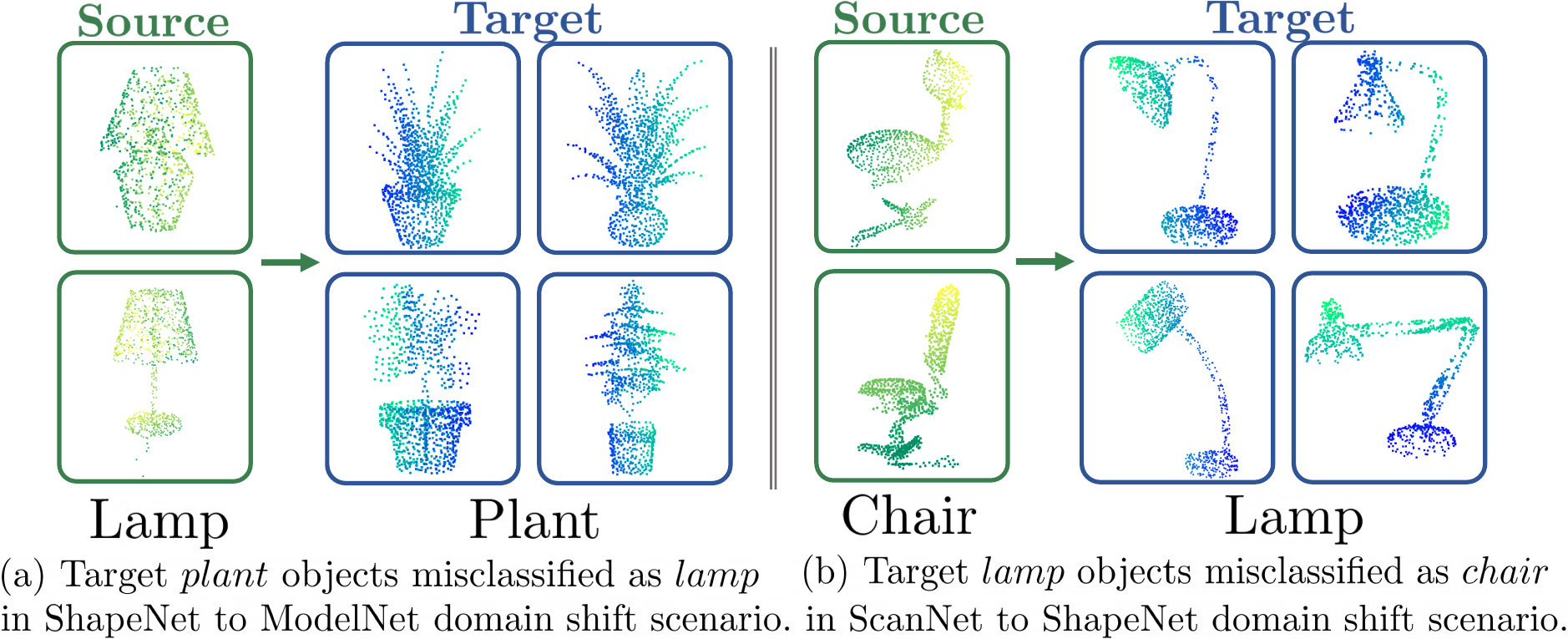}
   \label{fig:shapenet_to_modelnet_pt_misprediction}
  \vspace{-0.05in}
  \caption{\small{\textcolor{blue}{\textbf{Target}} point cloud samples that Point Transformer mispredicts as the ground-truth classes of \textcolor{teal}{\textbf{source}} samples on two domain shift scenarios (a) and (b). Regarding (a), the encoder perceives the vague outlines of target plants that resemble those of the source lamps, thereby mispredicting the target objects without comprehending the precise structure (leaves, for instance). A similar tendency is observed in (b), where round seats and curved outlines of source chairs are analogous to target lamps, resulting in the encoder misclassifying target lamps as chairs.}}
  \label{fig:pointtransformer_misprediction}
  \vspace{-0.1in}
\end{figure}

\begin{figure*}[t] 
  \centering
  \begin{minipage}[b]{0.86\textwidth} 
   \includegraphics[width=\textwidth]{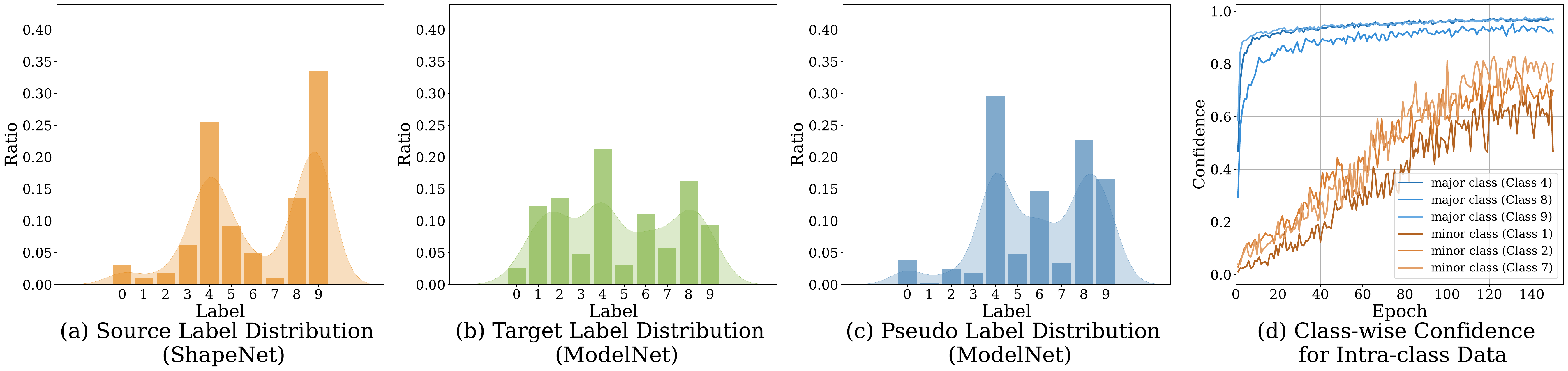}
  \end{minipage}
  \begin{minipage}[b]{0.86\textwidth} 
   \includegraphics[width=\textwidth]{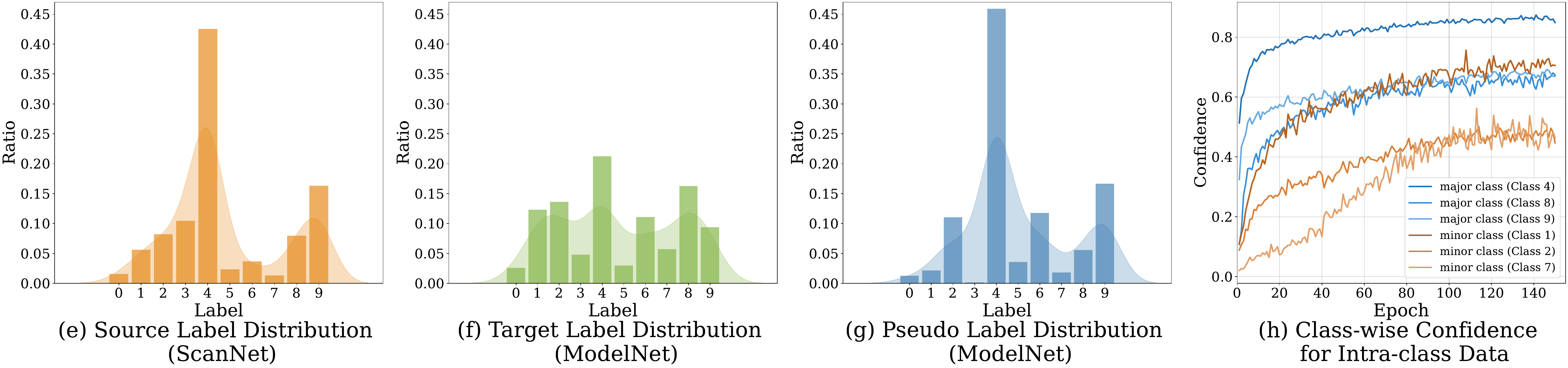}
  \end{minipage}
  \caption{\small{Label distributions (a-c and e-h) and class-wise confidence trends on source training data (d and h). The first row corresponds to ShapeNet to ModelNet domain shift scenario, whereas the below indicates the domain shift of ScanNet to ModelNet. According to the above figures, target pseudo-label distributions (c and g) tends to follow the source label distributions where source objects of minority classes show low confidence, injecting bias into the shared classifier and thereby rarely yielding pseudo-labels for the very classes.}}
  \label{fig:pl_distribution}
  \vspace{-0.05in}
\end{figure*}

Then the following question naturally arises: \textit{How to transfer the global anatomy information from the source domain to the target domain?} A naive approach for this distillation is leveraging a shared encoder and training it on both source and target domains. However, as existing point cloud encoding architectures~\cite{pointnet,pointnet++,dgcnn,pointtransformer,rscnn} are not specifically designed for capturing the global anatomy of the objects, they could often fail to identify this shape information. 
To confirm our claim,  we select the Point Transformer~\cite{pointtransformer},  one of the representative point cloud encoding architectures, as a shared encoder and test this model under two domain shift settings (ShapeNet-10$\rightarrow$ModelNet-10 and ScanNet-10$\rightarrow$ShapeNet-10). Then we visualize representative target objects that the encoder misclassifies in Figure~\ref{fig:pointtransformer_misprediction}. According to the figure, we discover that Point Transformer tends to predict target objects depending solely on the \textit{specific} part, without full consideration. From these observations, we could claim the following remark as the design principle of the domain adaptation framework: 
\begin{remark}\label{remark3.1}
Knowledge of global geometry is potentially conducive to domains with distributional shifts, whereas local features of point clouds need to be thoroughly adapted toward the target domain. To transfer the geometry information across domains, a sophisticated encoding module capturing objects in the entirety is necessary for adaptation.
\end{remark}
\vspace{-0.05in}
\begin{figure*}[t] 
  \centering
   \includegraphics[width=0.81\textwidth]{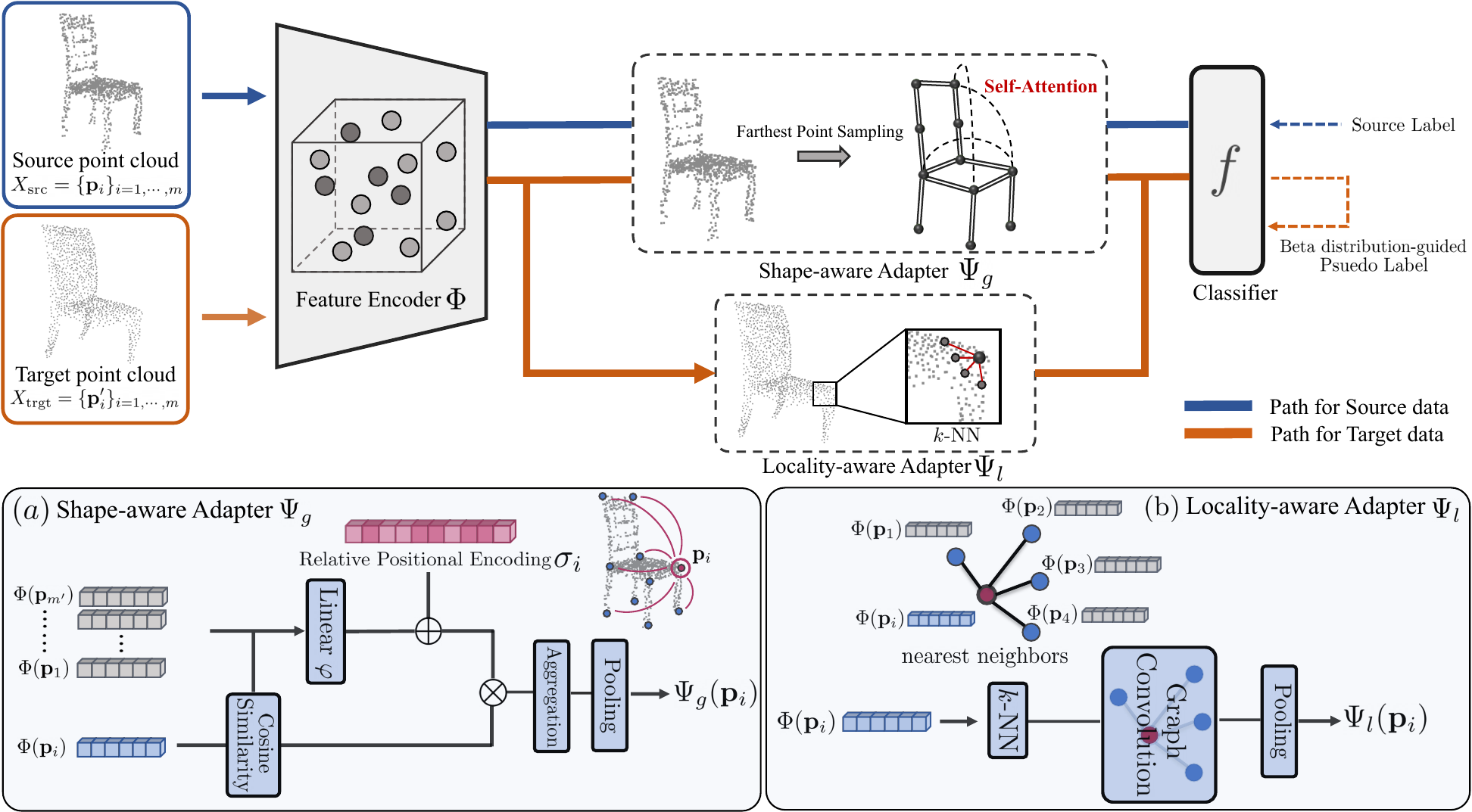}
  \caption{\small{Illustration of proposed framework \textit{PC-Adapter}. Our framework provides distinct training paths for different domains to conduct effective domain adaptation. To begin with, the \textit{PC-Adapter} feeds $\bm X_{\text{src}}$ to shared $\Phi$ and learns the global anatomy of the source domain from $\Psi_g$ that utilizes relative positional encoding $\sigma$ with farthest sampled points and $\Phi(\bm X_{\text{src}})$. Meanwhile, the pathway for $\bm X_{\text{trgt}}$ includes all key components $\Phi$, $\Psi_g$, and $\Psi_l$. Following $\Phi$, the proposed framework perceives global structures of $\bm X_{\text{trgt}}$ from $\Psi_g$ trained by the source domain. In conjunction with global knowledge adjustment, our \textit{PC-Adapter} learns local properties of $\bm X_{\text{trgt}}$ via $\Psi_l$ to compensate for the characteristic discrepancy resulting from the domain shift. Finally, the adapter outputs are combined for each domain and transmitted to the shared classifier $f$. In addition, \textit{PC-Adapter} leverages unbiased pseudo-labels for target point clouds with the guidance of beta distribution.}}
  \label{fig:concept_figure}
  \vspace{-0.05in}
\end{figure*}

\vspace{0.06in}
\subsection{Analysis of Pseudo-label Distribution}~\label{subsec:reamrk3.3}
Regarding another key ingredient of UDA for point clouds, pseudo-labeling, we investigate how the model trained on source labels affects the distribution of pseudo-labels obtained for target point clouds. Most recent domain adaptation works on point cloud data including \cite{glrv, gast, mlsp, shen2022domain} involve a \textit{self-training} technique, which utilizes pseudo-labels of target point clouds to train the model. Pseudo-labels are typically determined by the prediction scores of the classifier and are adopted only when the maximum confidence score is above a certain threshold $\gamma$. However, this strategy has a risk that the classifier could recklessly predict target point clouds as major classes of the source domain. To verify this, we explore pseudo-label distributions target training point clouds (Figure~\ref{fig:pl_distribution}). DGCNN models are trained under two domain shift settings (ShapeNet $\rightarrow$ ModelNet and ScanNet $\rightarrow$ ModelNet) using basic self-training where pseudo-labels for target data are determined by the classes of maximum confidence scores given a fixed threshold ($\gamma=0.9$). As shown in Figure~\ref{fig:pl_distribution} (c) and (g), pseudo-labels indeed follow the (ground-truth) label distribution of the source training data (Figure~\ref{fig:pl_distribution} (a) and (e)), and pseudo-labels are rarely selected for the minority classes at the source domain. Imposing such an inductive bias of source label distribution onto the arbitrary target domains is problematic as they could have far different label distributions.

While previous works have attempted to generate accurate pseudo-labels on domain-shifted data, these methods are insufficient to alleviate this problem as they merely focus on the reliability of pseudo-labels. Specifically, \cite{gast} increasingly selects pseudo-labels of target samples with low confidences  (\ie harder samples) to train a model in an easy-to-hard learning manner, and \cite{glrv} uses voting to select pseudo-labels based on the labels of nearest neighbors in the representation space to ensure the consistency of pseudo-labels. Although these methods can filter out noisy pseudo-labels to some extent by focusing on the \textit{magnitude} of maximum, they are not enough to prevent pseudo-labels from following the source label distribution as they cannot change the \textit{order of prediction scores} between classes. 
The reason for modifying the order of prediction scores can be found in our following observation. In Figure~\ref{fig:pl_distribution} (d) and (h), we measure the class-wise confidence on the source training data, computed by averaging the confidences on intra-class data. Despite confidences only for intra-class samples, we confirm that the model predicts the samples from the minor classes with low confidences during the training phase though they actually belong to corresponding minor classes. Hence, before generating pseudo-labels, rectifying the confidences considering the classifier bias is highly required to obtain unbiased pseudo-labels. Our second design principle for domain adaptation on point cloud data is as follows:
\begin{remark}~\label{remark3.2}
Pseudo-labels for target point clouds are readily biased toward the source label distribution. As classifier bias leads to varying scales of intra-class prediction probabilities for major/minor class samples, this phenomenon should be reflected in pseudo-label generation.
\end{remark}

\section{Proposed Method: PC-Adapter}
Building upon the design principles outlined in Section~\ref{sec:prelim_study}, we introduce \textit{PC-Adapter}, an effective domain adaption framework for 3D point clouds. Our method is composed of two key components. (1) To \textit{selectively} distill the object information from the source emphasized in Remark~\ref{remark3.1}, we employ two adapters ($\Psi_g$ and $\Psi_l$) inserted between the encoder $\Phi$ and classifier $f$ (Section~\ref{subsec:adap_da}). The intuition behind leveraging the adapter module for domain adaptation is that it can serve as a mediator, facilitating the transfer of versatile shape knowledge across domains via shape-aware adapter $\Psi_g$ while intensifying the adaptation to local characteristics of the target domain through target-exclusive locality-aware adapter $\Psi_l$. (2) Our \textit{PC-Adapter} generates qualified pseudo-labels for target point clouds that are resistant to classifier bias, as motivated in Remark~\ref{remark3.2}. Prior to pseudo-label selection, our method adjusts the prediction score in advance by referring to the relative percentile on its confidence distribution (Section~\ref{subsec:unbiased_pl}). The overall pipeline of \textit{PC-Adapter} is presented in Figure~\ref{fig:concept_figure} and the training algorithm in Appendix \textcolor{red}{C}. We describe the details of each component in the following subsections.

\subsection{Adapter-based Domain Adaptation}~\label{subsec:adap_da}
\vspace{-0.35in}
\paragraph{Shape-aware adapter}\label{global_adapter}
We now present the shape-aware adapter $\Psi_g$ to effectively encode the knowledge of global structure implied in both the source and target domains. To identify the shape of objects, we design our adapter to compute self-attention~\cite{vaswani2017attention} between the farthest points of an object (Figure~\ref{fig:concept_figure}). For a given 3D point cloud $\bm{X}_k$, a set of farthest points $\tilde{\bm{X}}_k = \{\mathbf{p}_i\}_{i=1}^{m'}$  are sampled from $\bm{X}_k$ using FPS algorithm. Then self-attention operations are conducted among the farthest points following attention weights $w_{ij}$, computed as cosine similarity between outputs of feature encoder $\Phi$:
\begin{equation}
\begin{aligned}
    \Psi_g(\mathbf{p}_i) =& \sum_{\mathbf{p}_j\in \tilde{\bm{X}_k}\backslash\mathbf p_i} w_{ij}(\varphi(\Phi(\mathbf{p}_j))+\sigma_{ij}),\\
     w_{ij} &= \frac{\Phi(\mathbf{p}_i)\cdot \Phi(\mathbf{p}_j)}{\|\Phi(\mathbf{p}_i)\|\|\Phi(\mathbf{p}_j)\|},
\end{aligned}
\end{equation}
where $\varphi$ denotes a linear projection layer,  
and $\sigma_{ij}$ represents a position encoding. Note here that the shape-aware adapter $\Psi_g$ consists of only \textit{one} self-attention layer.


As we have argued in Remark~\ref{subsec:remark3.1}, properly capturing shape information of the source and conveying this knowledge to the target domain facilities the interpretation of domain-shifted objects. To achieve this, we propose a novel relative positional encoding that identifies the global structure by considering both the relative point location and the ratio information of the objects.
Our positional encoding utilizes the relative distances from $\mathbf{p}_i$ to the remaining FPS points guided by the maximum distance concerning $\mathbf{p}_i$, $d_{i}^{*}=\max_{\scriptstyle{\mathbf{p}_j \in \tilde{\bm{X}}_k}}(\|\mathbf{p}_i - \mathbf{p_j}\|)$. Specifically, for a given point $\mathbf{p}_i \in \tilde{\bm{X}_k}$, the relative distance $\hat d_{ij}$ with another FPS point $\mathbf{p}_j$ is derived by subtracting the maximum distance in terms of $\mathbf{p}_i$, denoted as $d_{i}^{*}$, to the euclidean distance between $\mathbf{p}_i$ and $\mathbf{p}_j$, $d_{ij}=||\mathbf{p}_i-\mathbf{p}_j||$. Then, we obtain positional encoding $\sigma_{ij}$ by normalizing $\hat d_{ij}$ over other remaining relative distances $\{\hat d_{in} | \mathbf{p}_n \in \tilde{\bm{X}}_k \}$ to sum to 1 as follows: 
\begin{equation}
    \sigma_{ij} = \frac{\hat d_{ij}}{\sum_{n=1}^{m'} \hat d_{in}}
    = \frac{d_{i}^{*} - d_{ij}}{\sum_{n=1}^{m'} (d_{i}^{*}-d_{in})}.
\end{equation}
The above formulation allows us to yield $\mathbf{p}_i$-specific positional encoding values. As a result, our relative positional encoding approach using $d_{i}^{*}$ contemplates the object structure from diverse perspectives, enabling $\Psi_g$ to encode implicit global geometry in a more sophisticated fashion.


\paragraph{Locality-aware adapter}\label{local_adapter}
By incorporating the shape-aware adapter, the global shape knowledge would be effectively shared between source and target domains. Nevertheless, for successful domain adaptation, it is indispensable to introduce an exclusive module that can learn target-specialized local characteristics. To this end, we design a locality-aware adapter $\Psi_l$, which is founded on Graph Convolution layer~\cite{gcn} to aggregate regional information (Figure~\ref{fig:concept_figure} (b)). For a given point $\mathbf{p}_i$ sampled by FPS, \textit{k}-nearest neighbors (\textit{k}-NN) graph is constructed with nearest local neighbors \hjs{among full input points}, denoted as $\mathcal{N}(i)$. Then, the Graph Convolution operation is performed on the \textit{k}-NN graph to update the representation of $\mathbf{p}_i$ via adjacent point embeddings. This operation is formulated as follows:
\begin{equation}
\Psi_l(\mathbf{p}_i) = \mathbf{\Theta}^{\mathsf{T}}\sum_{\scriptstyle{\mathbf{p}_j\in \mathcal{N}(i)\cup\{\mathbf{p}_i\}}}\frac{e_{j,i}}{\texttt{deg}(\mathbf{p}_j)\texttt{deg}(\mathbf{p}_i)}\Phi(\mathbf{p}_j),
\end{equation}
where $\mathbf{\Theta}$ is a matrix of filter parameters, $e_{ji}$ is an edge weight of edge $\{\mathbf{p_j}, \mathbf{p_i}\}$, and $\texttt{deg}(\mathbf{p}_i)$ is a degree normalization constant computed as $\texttt{deg}(\mathbf{p}_i)=  1 + \sum_{\mathbf{p}_j \in \mathcal{N}(i)} {e_{j, i}}$. In this work, we consider edge feature $e_{ji}$ as 1. We would like to emphasize that locality-aware adapter $\Psi_l$ passes only the target data, generating representation vectors that are added to the existing model path via a residual connection. The detailed training procedures of ours are described at the following paragraph.

\paragraph{Training procedure}\label{training_procedure}
We here describe the training procedure of \textit{PC-Adapter} using a single point cloud instance per each domain for brevity, though the model is trained by mini-batches. 
To promote effective adaptation, we provide \textit{different} training paths for the source and target domains. For each iteration, we pass source data $\bm{X}^{\text{src}}$=$\ \{\mathbf{p}_i\}^{m}_{i=1}$ to the shared encoder $\Phi$, and the encoder output $\{\Phi(\mathbf{p}_i)\}^{m}_{i=1}$ with its coordinate information is fed to the shape-aware adapter $\Psi_g$. Then, $\Psi_g$ learns the global shape of the source domain by performing self-attention between representative points. The representation obtained from $\Psi_g$ is combined with the original encoder output, $\texttt{Combine} \big(\{\Psi_g(\mathbf{p}_i)\}_{i=1}^{m'},\{\Phi(\mathbf{p}_i)\}_{i=1}^{m}\big)$, and then transmitted to the classifier $f$.  
 Subsequently, our method makes the target data $\bm{X}^{\text{trgt}}$=$\ \{\mathbf{p}'_i\}^{m}_{i=1}$ forward through all components, \ie $\Phi$, $\Psi_g$, and $\Psi_l$. Following the shared $\Psi_g$, \textit{PC-Adapter} identifies the global anatomy of $\bm{X}^{\text{trgt}}$ by receiving learned global knowledge of the source.
In conjunction with global structure adaptation, our \textit{PC-Adapter} learns local properties via locality-aware adapter $\Psi_l$ to compensate for the characteristic discrepancy between two domains. Lastly, the outputs of two adapters are combined with encoder output as $\texttt{Combine} \big(\{\Psi_g(\mathbf{p}_i)\}_{i=1}^{m'}, \{\Psi_l(\mathbf{p}_i)\}_{i=1}^{m'}, \{\Phi(\mathbf{p}_i)\}_{i=1}^{m'}\big)$, and they are fed into the classifier $f$. During training target point clouds, we lower the learning rate for components where the source data have passed (\ie $\Phi$ and $\Psi_g$) to preserve the implicit shape knowledge of the source domain.
The efficacy of weakly updating these components is confirmed in Appendix \textcolor{red}{A}. We select the \texttt{Combine} operation among summation and averaging.
\begin{table*}[t]
\caption{\small Classification accuracy of proposed \textit{PC-Adapter} against other baselines on PointDA-10, averaged over three repetitions ($\pm$ SEM). \textit{SSL}, \textit{Adv}., and \textit{ST} denote self-supervised learning, adversarial learning, and self-training (include data augmentation), respectively.} 
\vspace{-0.05in}
\begin{center}
\begin{footnotesize}
\setlength{\columnsep}{1pt}%
\begin{adjustbox}{width=0.89\linewidth}
\begin{tabular}{@{\extracolsep{1pt}}lccc|cccccc|l@{}}
\toprule
 \textbf{Method} & \textit{SSL}& \textit{Adv.} & \textit{ST} & M$\rightarrow$S & M$\rightarrow$S* & S$\rightarrow$M & S$\rightarrow$S* & S*$\rightarrow$M & S*$\rightarrow$S & Avg.  \\ 
\cline{1-11}
                    Supervised & & & &  93.9 \tiny{$\pm 0.2$} & 78.4 \tiny{$\pm 0.6$} & 96.2 \tiny{$\pm 0.1$} & 78.4 \tiny{$\pm 0.6$} & 96.2 \tiny{$\pm 0.1$} & 93.9 \tiny{$\pm 0.2$} & 89.5 \\
                    \cdashline{1-11}
                    DANN~\cite{dann} & & \cmark &  & 74.8 \tiny{$\pm 2.8$} & 42.1 \tiny{$\pm 0.6$} & 57.5 \tiny{$\pm 0.4$} & 50.9 \tiny{$\pm 1.0$} & 43.7 \tiny{$\pm 2.9$} & 71.6 \tiny{$\pm 1.0$} & 56.8 \\
                    PointDAN~\cite{pointdan} & & \cmark & & \textbf{83.9 \tiny{$\pm 0.3$}} & 44.8 \tiny{$\pm 1.4$} & 63.3  \tiny{$\pm 1.1$} & 45.7 \tiny{$\pm 0.7$} & 43.6 \tiny{$\pm 2.0$} & 56.4 \tiny{$\pm 1.5$} & 56.3 \\
                    DefRec+PCM~\cite{defrec} & \graycheck & & \cmark & 83.7 \tiny{$\pm 0.6$} & 42.6 \tiny{$\pm 0.9$} & 71.4 \tiny{$\pm 1.5$} & 46.1 \tiny{$\pm 1.7$} & 71.5 \tiny{$\pm 1.0$} & 74.6 \tiny{$\pm 0.5$} & 65.0 \\
                    GAST~\cite{gast}  & \graycheck & & \cmark & 83.7 \tiny{$\pm 0.2$} & 54.9 \tiny{$\pm 0.9$} & 71.1 \tiny{$\pm 0.2$} & 53.6 \tiny{$\pm 0.2$} & 48.4 \tiny{$\pm 3.3$} & 58.7 \tiny{$\pm 0.1$} & 61.7 \\
                    GLRV~\cite{glrv}  & \graycheck & \cmark & \cmark & 78.4 \tiny{$\pm 0.6$} & 53.6 \tiny{$\pm 0.1$} & 64.4 \tiny{$\pm 3.8$} & 49.1 \tiny{$\pm 2.8$} & 59.4 \tiny{$\pm 2.5$} & 71.2 \tiny{$\pm 3.7$} & 62.7 \\
                    \cline{1-11}
                    \textbf{\textit{PC-Adapter}} & & & \cmark & 83.3 \tiny{$\pm 0.3$} & \textbf{58.2} \tiny{$\pm 0.4$} & \textbf{77.5} \tiny{$\pm 0.2$} & \textbf{53.7} \tiny{$\pm 0.2$} & \textbf{73.7} \tiny{$\pm 0.5$} & \textbf{75.4} \tiny{$\pm 0.4$} & \textbf{70.3 }\\

 \bottomrule
\end{tabular}
\end{adjustbox}
\end{footnotesize}
\end{center}
\label{tb:main_pointDA}
\vspace{-0.05in}
\end{table*}

\begin{table*}[t]
\caption{\small Classification accuracy of the \textit{PC-Adapter} compared to baselines on GraspNetPC-10, averaged over three repetitions ($\pm$ SEM). Syn., Kin., and RS. refer to the synthetic, kinetic, and realsense domains, respectively.} 
\vspace{-0.05in}
\begin{center}
\setlength{\columnsep}{1.2pt}%
\begin{adjustbox}{width=0.75\linewidth}
\begin{tabular}{@{\extracolsep{1pt}}lccc|cccc|l@{}}
\toprule
  \textbf{Method} & \textit{SSL}& \textit{Adv.} & \textit{ST} & Syn.$\rightarrow$Kin. & Syn.$\rightarrow$RS. & Kin.$\rightarrow$RS. & RS.$\rightarrow$Kin. & Avg.  \\ 
\cline{1-9}
                    Supervised & & & & 97.2 \tiny{$\pm 0.8$} & 95.6 \tiny{$\pm 0.4$} & 95.6 \tiny{$\pm 0.3$} & 97.2 \tiny{$\pm 0.4$} & 96.4 \\
                    \cdashline{1-9}
                     DANN~\cite{dann} & & \cmark & & 78.6 \tiny{$\pm 0.3$} & 70.3 \tiny{$\pm  0.5$} & 46.1  \tiny{$\pm 2.2$} & 67.9 \tiny{$\pm 0.3$} & 65.7  \\
                     PointDAN~\cite{pointdan} & & \cmark & & 77.0 \tiny{$\pm 0.2$} & 72.5 \tiny{$\pm 0.3$} & 65.9  \tiny{$\pm 1.2$} & 82.3 \tiny{$\pm 0.5$} & 74.4 \\
                     RS~\cite{rs} & \cmark & & & 67.3 \tiny{$\pm 0.4$} & 58.6 \tiny{$\pm 0.8$} & 55.7 \tiny{$\pm 1.5$} & 69.6 \tiny{$\pm 0.4$} & 62.8 \\
                     DefRec+PCM~\cite{defrec} & \cmark & & \cmark & 80.7 \tiny{$\pm 0.1$} & 70.5 \tiny{$\pm 0.4$} &  65.1  \tiny{$\pm 0.3$} & 77.7 \tiny{$\pm 1.2$} & 73.5 \\
                     GAST~\cite{gast} & \cmark & & \cmark & 81.3 \tiny{$\pm  1.8$} & 72.3 \tiny{$\pm 0.8$} & 61.3  \tiny{$\pm 0.9$} &  80.1 \tiny{$\pm 0.3$} & 73.8 \\
                     GLRV~\cite{glrv} & \cmark & \cmark & \cmark & 92.2 \tiny{$\pm  2.1$} & 74.4 \tiny{$\pm 0.8$} & 70.2  \tiny{$\pm 2.4$} &  86.7 \tiny{$\pm 0.3$} & 73.8 \\
                     ImplicitPCDA~\cite{shen2022domain} & \cmark & & \cmark & 94.6  \tiny{$\pm  0.4$} & 80.5 \tiny{$\pm 0.2$} & 76.8 \tiny{$\pm 0.4$} &  85.9 \tiny{$\pm 0.3$} & 84.4 \\
                     \cline{1-9}
                     \textbf{\textit{PC-Adapter}+ImplicitPCDA} & \cmark & & \cmark & \textbf{97.5} \tiny{$\pm 0.3$} & 72.5 \tiny{$\pm 0.2$} & \textbf{82.5} \tiny{$\pm 0.7$} & \textbf{88.6} \tiny{$\pm 1.7$} & \textbf{85.3} \\
                    
\bottomrule
\end{tabular}
\end{adjustbox}
\end{center}
\label{tb:main_graspnet}
\vspace{-0.1in}
\end{table*}

\subsection{Distribution-guided Unbiased Pseudo-label}~\label{subsec:unbiased_pl}
There remains a challenge of how to annotate target point clouds with pseudo-labels which are readily biased towards the source label distribution. With this goal in mind, we devise a bias-resilient pseudo-labeling method that identifies the classifier bias in advance and adjusts model confidences before selecting pseudo-labels. 
We start by approximating the confidence distribution for each class $t$ given source training dataset $\mathcal{S}_\text{train}$ as a skewed beta distribution, $p(c_t|\mathcal{S}_{\text{train}}) \approx \texttt{Beta}(\hat{\alpha}_t, \hat{\beta}_t)$. 
We justify this modeling for the following two reasons: (1) the average prediction confidence for intra-class data of each class is significantly different, which strongly correlates with the source label distribution (Figure~\ref{fig:pl_distribution} (d) and (h)). (2) the skewness of confidence distribution is determined by the degree of a majority of the class. In other words, for major classes, confidence distributions are likely to be left-tailed as samples of major classes are frequently encountered during training, while for minor classes, the confidence distributions are more likely to be right-tailed. Let us denote the indices of training samples belonging to class $t$ as $\mathcal{S}^t_{\text{train}}$, and the confidence on class $t$ for the $i$-th sample as $c_{i,t}$. Then, for each class $t$, two unknown parameters - $\hat{\alpha}_t$ and $\hat{\beta}_t$ - can be estimated using sample mean of confidences, $\bar{c}_t = \frac{1}{|\mathcal{S}^{t}_{\text{train}}|}\sum_{i\in \mathcal{S}^{t}_{\text{train}}}c_{i,t}$, and sample variance, $\bar{v}_t = \frac{1}{|\mathcal{S}^{t}_{\text{train}}|-1}\sum_{i\in \mathcal{S}^{t}_{\text{train}}}(c_{i,t}-\bar{c}_t)^2 $, based on Method of Moments~\cite{mom}: 
\begin{equation}
\hat{\alpha}_{t} = \bar{c}_t\big(\frac{\bar{c}_t(1-\bar{c}_t)}{\bar{v}_t}-1\big), \ \hat{\beta}_{t} = (1-\bar{c}_t)\big(\frac{\bar{c}_t(1-\bar{c}_t)}{\bar{v}_t}-1\big).  
\end{equation}The detailed proof is provided at Appendix \textcolor{red}{B}.

Armed with approximated class-wise confidence distributions, our method adjusts the confidence score of each class by reflecting its percentile rank within its own confidence distribution, prior to assigning pseudo-labels. By considering the \textit{relative} confidence of each class within its distribution, our method encourages the selection of minority classes as pseudo-labels, thereby allowing the model to obtain unbiased pseudo-labels without requiring access to the label distribution of the target domain. Specifically, our pseudo-labeling method modifies the confidence score $c_{i,t}$ of each class $t$ such that it increases proportionally with the percentile rank $r_i$ as follows: $\tilde{c}_{i,t}=c_{i,t}\cdot(\frac{1}{1-r_i + r_0})$ to remove the effect of bias in the source label distribution, where $r_i$ is derived from percentile point function of beta distribution (\ie inverse cumulative density function) and $r_0$ is a hyperparameter that controls the intensity of adjustment. Lastly, the pseudo-label $\hat{y}_{i}^{\text{trgt}}$ for the target point cloud $\bm{X}^{\text{trgt}}_{i}$ is annotated as ${\arg \max}_{t}\ \tilde{c}_{i,t}$. Note that our pseudo-labeling strategy requires no additional model forward/backward pass as distributions are computed based on the model predictions acquired during the training phase.

\jh{While our distribution-guided pseudo-labeling mitigates the selection of biased pseudo-labels for target data, in extreme cases (\eg when the source domain exhibits high class imbalance), our confidence rectification might not succeed in altering the order of prediction scores due to significant gaps between intra-class prediction scores. To address this challenge, we incorporate the regularization loss term $\mathcal{L}_{\text{centroid}}$ for each training batch, enforcing the orthogonalization of averaged representations for each class.}

\section{Experiments}
\jh{We conduct a series of experiments to verify our \textit{PC-Adapter} in various domain shift scenarios. In Section~\ref{subsec:main_classification} and \ref{subsec:main_segmentation}, we evaluate the adaptation performance on two shape classification benchmarks and one part segmentation benchmark dataset, comparing it with recent unsupervised domain adaptation methods for the point cloud. To further demonstrate the effectiveness of our framework, we test our method under data-scarce conditions and analyze its time complexity (Section~\ref{subsec:further_analysis}). Lastly, we perform an ablation study to examine the contribution of each sub-component of \textit{PC-Adapter} (Section~\ref{subsec:ablation_study}).}


\begin{table*}[t]
\caption{\small Part segmentation results on the PointSegDA dataset. Mean per-class IoU (mIoU) for our method and other baselines are reported, averaged over three repetitions ($\pm$ SEM).}
\begin{center}
\setlength{\columnsep}{1pt}%
\begin{adjustbox}{width=\linewidth}
\begin{tabular}{@{\extracolsep{1pt}}lcccccccccccc|l@{}}
\toprule
 \textbf{Method} & \shortstack{FAUST$\rightarrow$\\ADOBE} & \shortstack{FAUST$\rightarrow$\\MIT}  & \shortstack{FAUST$\rightarrow$\\SCAPE}  & \shortstack{MIT$\rightarrow$\\ADOBE}  & \shortstack{MIT$\rightarrow$\\FAUST}  & \shortstack{MIT$\rightarrow$\\SCAPE}  & \shortstack{ADOBE$\rightarrow$\\FAUST}  & \shortstack{ADOBE$\rightarrow$\\MIT}  & \shortstack{ADOBE$\rightarrow$\\SCAPE}  & \shortstack{SCAPE$\rightarrow$\\ADOBE}  & \shortstack{SCAPE$\rightarrow$\\FAUST}  & \shortstack{SCAPE$\rightarrow$\\MIT}  & Avg.  \\ 
\cline{1-14}
                    Supervised & 80.9 \tiny{$\pm 7.2$} & 81.8 \tiny{$\pm 0.3$} & 82.4 \tiny{$\pm 1.2$} & 80.9 \tiny{$\pm 7.2$} & 84.0 \tiny{$\pm 1.8$} & 82.4 \tiny{$\pm 1.2$} & 84.0 \tiny{$\pm 1.8$} &  81.8 \tiny{$\pm 0.3$} & 82.4 \tiny{$\pm 1.2$} & 80.9 \tiny{$\pm 7.2$} & 84.0 \tiny{$\pm 1.8$} & 81.8 \tiny{$\pm 0.3$} & 82.3  \\

                    \cdashline{1-14}
                     Unsupervised & 78.5 \tiny{$\pm 0.4$} & 60.9 \tiny{$\pm 0.6$} & 66.5  \tiny{$\pm 0.6$} & 26.6 \tiny{$\pm 3.5$} & 33.6 \tiny{$\pm 1.3$} & 69.9 \tiny{$\pm 1.2$} & 38.5 \tiny{$\pm 2.2$} & 31.2 \tiny{$\pm 1.4$} & 30.0 \tiny{$\pm 3.6$} & 74.1 \tiny{$\pm 1.0$} & 68.4 \tiny{$\pm 2.4$} & 64.5 \tiny{$\pm 0.5$} & 53.6 \\
                     Adapt-SegMap~\cite{adapt_segmap} & 70.5 \tiny{$\pm 3.4$} &60.1 \tiny{$\pm 0.6$}& 65.3 \tiny{$\pm 1.3$}& 49.1 \tiny{$\pm 9.7$}& 54.0 \tiny{$\pm 0.5$} &62.8 \tiny{$\pm 7.6$}& 44.2 \tiny{$\pm 1.7$} &35.4 \tiny{$\pm 0.3$} &35.1 \tiny{$\pm 1.4$}& 70.1 \tiny{$\pm 2.5$} & 67.7 \tiny{$\pm 1.4$} & 63.8 \tiny{$\pm 1.2$} &56.5 \\
                     RS~\cite{rs} & 78.7 \tiny{$\pm 0.5$} & 60.7 \tiny{$\pm 0.4$} & 66.9 \tiny{$\pm 0.4$} & 59.6 \tiny{$\pm 5.0$} & 38.4 \tiny{$\pm 2.1$} & 70.4 \tiny{$\pm 1.0$} & 44.0 \tiny{$\pm 0.6$} & 30.4 \tiny{$\pm 0.5$} & 36.6 \tiny{$\pm 0.8$} & 70.7 \tiny{$\pm 0.8$} & 73.0 \tiny{$\pm 1.5$} & 65.3 \tiny{$\pm 0.1$} & 57.9 \\
                     DefRec+PCM~\cite{defrec} & 79.7 \tiny{$\pm 0.3$} & 61.8 \tiny{$\pm 0.1$}  &67.4 \tiny{$\pm 1.0$} & 67.1 \tiny{$\pm 1.0$} & 48.6 \tiny{$\pm 2.4$} & 72.6 \tiny{$\pm 0.5$} & 46.9 \tiny{$\pm 1.0$} & 33.2 \tiny{$\pm 0.3$} & 37.6 \tiny{$\pm 0.1$} & 66.4 \tiny{$\pm 0.9$} & 72.2 \tiny{$\pm 1.2$} & 66.2 \tiny{$\pm 0.9 $} & 60.0 \\
                     \cline{1-14}
                     \textbf{\textit{PC-Adapter}} & 78.8 \tiny{$\pm 0.5$} & 61.5 \tiny{$\pm 0.1$} & \textbf{69.2} \tiny{$\pm 0.3$} & \textbf{68.9} \tiny{$\pm 0.2$} & \textbf{62.1} \tiny{$\pm 0.3$} & \textbf{72.8} \tiny{$\pm 0.1$}  & \textbf{49.0} \tiny{$\pm 0.1$} & \textbf{39.6} \tiny{$\pm 0.1$}  & \textbf{43.1} \tiny{$\pm 0.1$} &\textbf{74.4} \tiny{$\pm 0.7$} &\textbf{75.7} \tiny{$\pm 0.3$} &\textbf{66.4} \tiny{$\pm 0.1$} & \textbf{63.5} \\
                    
\bottomrule
\end{tabular}
\end{adjustbox}
\end{center}
\label{tb:main_pointsegda}
\vspace{-0.1in}
\end{table*}

\subsection{Experimental Settings}
\paragraph{Datasets} Two \jh{shape classification} datasets, PointDA-10~\cite{pointdan} and GraspNetPC-10~\cite{shen2022domain} are adopted for our experiments. PointDA-10 is constructed by extracting the samples in 10 shared classes from three datasets - ModelNet40 (\textbf{M})~\cite{wu20153d}, ShapeNet (\textbf{S}) ~\cite{chang2015shapenet}, and ScanNet (\textbf{S*})~\cite{dai2017scannet}. Using these subsets, we conduct experiments on 6 different domain adaptation settings. To further verify our method on synthetic-to-real and real-to-real adaptation scenarios, we also conduct experiments on GraspNetPC-10~\cite{shen2022domain}. GraspNetPC-10 is curated via raw depth scans on real-world and also synthetic scenes which consist of 3D CAD models of objects using two different cameras, Kinect2 and Intel Realsense. Following \cite{shen2022domain}, we conduct experiments on 4 domain adaptation scenarios, where all domains share the same 10 classes. Among them, scenarios whose source domains correspond to Syn. represent \textit{synthetic}-to-real shifts, whereas the rest indicate real-to-real shifts where sensor noise differs. \jh{For 3D part segmentation tasks, we utilize PointSegDA dataset~\cite{defrec} in our experiments, which comprises 4 different subsets: FAUST, MIT, ADOBE, and SCAPE. These subsets encompass 8 classes of human body parts varying in point distribution, pose, and scanned individuals. We validate our framework on 12 different domain shift conditions, following the experimental settings of the \cite{defrec}.} Detailed statistics are provided in Appendix \textcolor{red}{D}.

\paragraph{Implementation detail}
In our experiments, we set the farthest point sampling (FPS) ratio to 0.1, and $k$ to 5 for $k$-NN graph in locality-aware adapter $\Psi_l$. We adopt DGCNN~\cite{dgcnn} as 
the feature encoder, consistent with prior works~\cite{gast,glrv,shen2022domain}. We follow the evaluation protocol of \cite{glrv,gast} for PointDA-10 dataset and \cite{shen2022domain} for GraspNetPC-10 dataset. \jh{For part segmentation experiments, we adhere to the experimental setup of \cite{defrec}.} Detailed experimental settings and more implementation details of ours are provided in Appendix \textcolor{red}{D}.

\subsection{Results on Shape Classification}~\label{subsec:main_classification}
 In Table~\ref{tb:main_pointDA}, we report the averaged classification accuracy of \textit{PC-Adapter} and baselines on 6 domain adaptation settings in PointDA-10. We first compare with UDA baselines that do not leverage auxiliary tasks (\ie self-supervised learning tasks) as our \textit{PC-Adapter} does. These baselines perform domain adaptation using such as adversarial domain alignment~\cite{pointdan, glrv,dann}, self-training~\cite{gast,glrv}, and mixup augmentation~\cite{defrec}. For this comparison, we exclude the self-supervised tasks from the methods that originally use them. As shown in Table~\ref{tb:main_pointDA}, our framework shows significant superiority over the baselines in most settings. The \textit{PC-Adapter} enhances the adaptation performance by an average of \textbf{5.3\%} compared to the best baseline results. For GraspNetPC-10, the \textit{PC-Adapter} exhibits state-of-the-art performances on three settings when combined with existing self-supervised learning tasks of ImplicitPCDA~\cite{shen2022domain} (Table~\ref{tb:main_graspnet}).  In particular, \textit{PC-Adapter} solely excels the performance over $80\%$ on Kin.$\rightarrow$ RS, yielding the improvement by \textbf{5.7}\% compared to the second highest result. From these results, we believe that our method can seamlessly integrate with existing self-supervised tasks and potentially exhibit superior performance, despite being designed for succinct adaptation-oriented purposes. 
 


\subsection{Results on Part Segmentation}~\label{subsec:main_segmentation}
\jh{While \textit{PC-Adapter} is primarily designed for point cloud classification tasks, our framework holds the potential for extension into various other 3D understanding tasks. To demonstrate the versatility of our approach, we perform experiments on the part segmentation tasks across 12 domain adaptation settings using the PointSegDA dataset. Unlike shape classification tasks, the crux of part segmentation lies in grasping local characteristics within each segment, enabling accurate predictions for individual points. In light of this, for part segmentation tasks, we only employ the locality-aware adapter $\Psi_l$ in the domain adaptation process, facilitating the transfer of knowledge concerning local characteristics from the source domain. When training target points, we simply annotate target points using pseudo-labels with a fixed threshold $\gamma$, while also reducing the learning rate for the feature encoder $\Phi$ and the locality-aware adapter $\Psi_l$, similar to that in shape classification tasks. In Table~\ref{tb:main_pointsegda}, our method achieves the best segmentation performances in 10 out of 12 domain shift settings, exhibiting an average mIoU enhancement of \textbf{3.5\%}.}

\begin{figure}[t] 
\centering
   \includegraphics[width=0.48\textwidth]{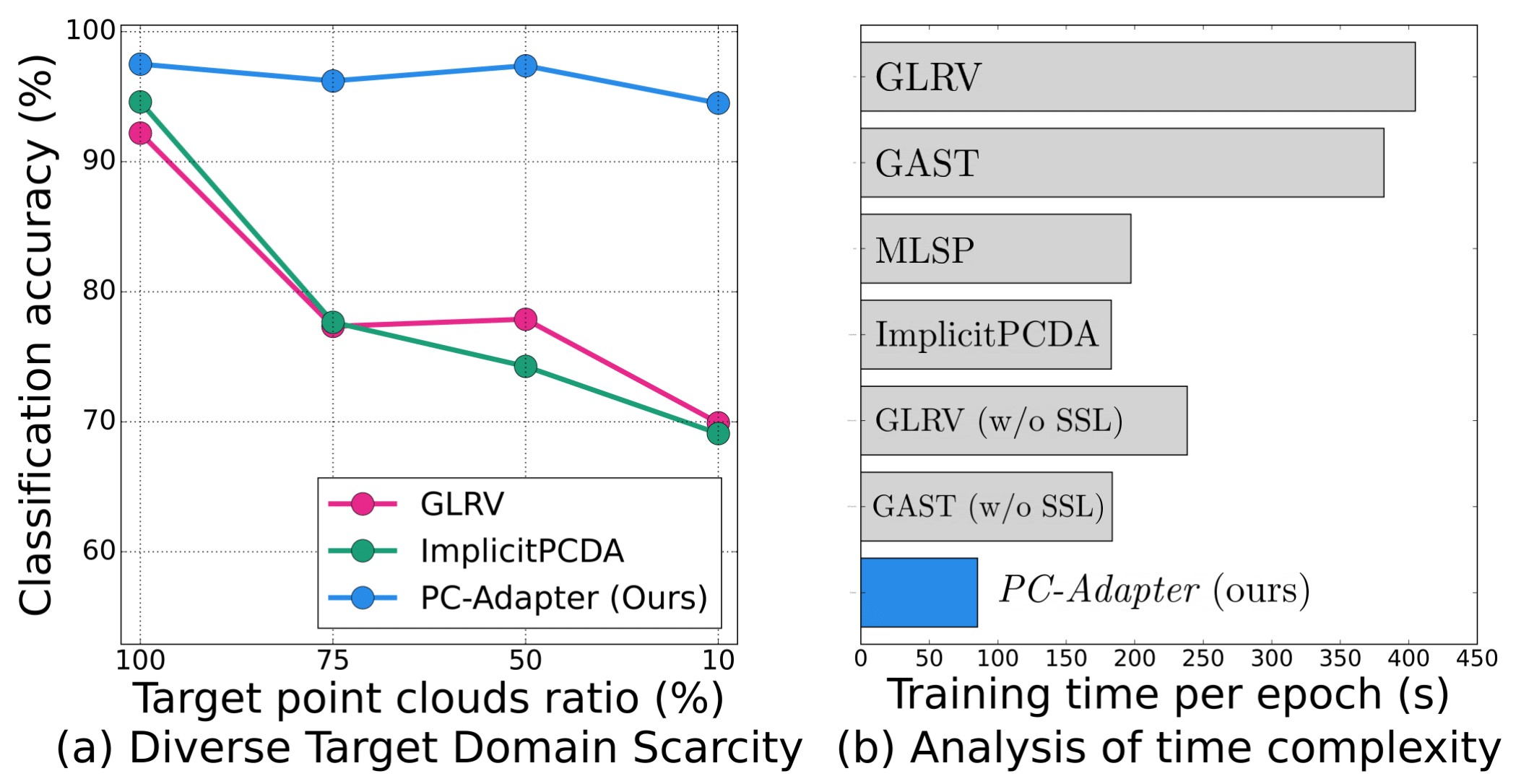}
  \caption{\small{(a) Classification accuracy under diverse target domain scarcity on Syn.$\rightarrow$Kin. (b) Time complexity analysis of the \textit{PC-Adapter} against baselines.}}
  \label{fig:further_analysis}
  \vspace{-0.1in}
\end{figure}

\subsection{Further Analysis}~\label{subsec:further_analysis}
\vspace{-0.2 in}
\paragraph{Data scarcity}
In practical scenarios, it is common to encounter situations where the target point clouds obtained from the real-world exhibit deficiencies when compared to synthetic source point clouds. To verify our method under this condition, we conduct an experiment by decreasing the number of target data to $75\%$, $50\%$, and $10\%$ at the synthetic-to-real domain shift setting in GraspNetPC-10. In Figure~\ref{fig:further_analysis} (a), the \textit{PC-Adapter} consistently shows the highest adaptation performances over other baselines across multiple sparsities. These results imply that, unlike baselines, our method can efficiently learn target-specific features using only a few target data is given.

\paragraph{Time Complexity}
To test the cost efficiency of our method, we compare the computation cost of \textit{PC-Adapter} with that of baselines. Training time per epoch is measured over 10 repetitions using an RTX 3090 GPU in the ModelNet$\rightarrow$ShapeNet setting. As shown in Figure~\ref{fig:further_analysis} (b), our method achieves \textbf{3.43x} faster training on average compared to the self-supervised learning works, and even faster than self-training (GAST \textit{w/o ssl}) and adversarial training (GLRV \textit{w/o ssl}) methods. This result demonstrates the advantage of deploying \textit{PC-Adapter} for real-world usage.

\subsection{Ablation Study}~\label{subsec:ablation_study}
\vspace{-0.2 in}
\paragraph{Step-wise Evaluation}
In Table~\ref{tb:stepwise_evaluation}, we conduct a step-wise evaluation to verify the significance of each component in \textit{PC-Adapter}:
(1) Domain adaptation using only a shape-aware adapter $\Psi_g$ (the first row), (2) Leveraging both adapters with basic pseudo-labeling (the second row), and (3) Replacing our relative positional encoding $\sigma$ with conventional point cloud positional encoding method~\cite{pointtransformer,rscnn}, formulated as $\sigma_{ij} = \theta(\mathbf{p}_i-\mathbf{p}_j)$, (the third row). The results clearly show that each module consistently improves performance under domain shift scenarios. 
\begin{table}[h]
\caption{\small Ablation study of \textit{PC-Adapter} on three domain shift settings in PointDA-10.} 
\vspace{-0.05in}
\begin{center}
\begin{footnotesize}
\setlength{\columnsep}{1pt}%
\begin{adjustbox}{width=0.84\linewidth}
\begin{tabular}{@{\extracolsep{1pt}}cccc|ccc}
\toprule
\multicolumn{4}{c}{\textbf{Modules}} & \multirow{1}{*}{S$\rightarrow$M} & \multirow{1}{*}{M$\rightarrow$S} & \multirow{1}{*}{S*$\rightarrow$M} \\
\cline{1-7}
 $\Psi_g$ & $\Psi_l$ & Beta PL & Relative $\sigma$ & Acc. & Acc.  & Acc. \\ 
\cline{1-7}
                    \cmark & & & \cmark & 72.3  & 81.4 & 67.4   \\
                    \cmark & \cmark & & \cmark & 75.7  & 82.7 & 68.2  \\
                    \cmark & \cmark & \cmark & & 74.2  & 81.5 &70.3 \\
                    \cmark & \cmark & \cmark & \cmark & \textbf{77.5} & \textbf{83.3} & \textbf{73.7} \\
\bottomrule
\end{tabular}
\end{adjustbox}
\end{footnotesize}
\end{center}
\label{tb:stepwise_evaluation}
\vspace{-0.1in}
\end{table}
\vspace{-0.1in}

\paragraph{Analysis of distribution-guided pseudo-labels} 
We quantitatively analyze the contribution of our pseudo-labeling method in comparison to maximum confidence-based pseudo-labels generation in Table~\ref{tb:bAcc}. Our beta distribution-guided pseudo-labeling consistently brings improvement in the balanced accuracy compared to the conventional pseudo-labeling approach, showing robustness to the label bias induced from the source domain.

\begin{table}[h]
\caption{\small Comparison of balanced accuracy (bAcc.) according to pseudo labeling strategies on GraspNetPC-10.}
\vspace{-0.05in}
\begin{center}
\setlength{\columnsep}{1pt}%
\begin{adjustbox}{width=\linewidth}
\begin{tabular}{@{\extracolsep{1pt}} l c c c c}
\toprule
\textbf{PL Methods} & Syn.$\rightarrow$Kin. & Syn.$\rightarrow$RS. & Kin.$\rightarrow$RS. & RS.$\rightarrow$Kin. \\
\hline
Maximum confidence PL & 96.41 & 67.93 & 79.49 & 87.93 \\
Beta distribution-guided PL (Ours) & \textbf{97.66} & \textbf{71.99} & \textbf{83.52} & \textbf{90.31} \\
\bottomrule
\end{tabular}
\end{adjustbox}
\end{center}
\label{tb:bAcc}
\vspace{-0.1in}
\end{table}

\section{Conclusion}
We scrutinized the essential design principles for domain adaption on point clouds, focusing on which information should be transferred during adaptation. Inspired by this, we proposed an efficient adapter-based domain adaptation framework, \textit{PC-Adapter}, that learns target-specific feature transformations while selectively preserving pertinent knowledge from the source. \textit{PC-Adpater} exhibits fast yet successful domain adaptation abilities under both normal and data-scarce conditions through extensive experiments.

\section*{Acknowledgements}
This work was supported by Institute of Information \& communications Technology Planning \& Evaluation (IITP) grant funded by the Korea government(MSIT) (No.2019-0-00075, Artificial Intelligence Graduate School Program (KAIST), No.2022-0-00713, Meta-learning applicable to real-world problems, No.2022-0-00984, Development of Artificial Intelligence Technology for Personalized Plug-and-Play Explanation and Verification of Explanation, No.2021-0-02068, Artificial Intelligence Innovation Hub) and the National Research Foundation of Korea (NRF) grants
(No.2018R1A5A1059921) funded by the Korea government
(MSIT). This work was also supported by Samsung Electronics Co., Ltd (No.IO201214-08133-01).
\clearpage
{\small
\bibliographystyle{ieee_fullname}
\bibliography{iccv2023}

\begin{thebibliography}{10}\itemsep=-1pt

\bibitem{defrec}
Idan Achituve, Haggai Maron, and Gal Chechik.
\newblock Self-supervised learning for domain adaptation on point clouds.
\newblock In {\em Proceedings of the IEEE/CVF winter conference on applications of computer vision}, pages 123--133, 2021.

\bibitem{chang2015shapenet}
Angel~X Chang, Thomas Funkhouser, Leonidas Guibas, Pat Hanrahan, Qixing Huang, Zimo Li, Silvio Savarese, Manolis Savva, Shuran Song, Hao Su, et~al.
\newblock Shapenet: An information-rich 3d model repository.
\newblock {\em arXiv preprint arXiv:1512.03012}, 2015.

\bibitem{vision_adapter}
Zhe Chen, Yuchen Duan, Wenhai Wang, Junjun He, Tong Lu, Jifeng Dai, and Yu Qiao.
\newblock Vision transformer adapter for dense predictions.
\newblock In {\em The Eleventh International Conference on Learning Representations}, 2023.

\bibitem{dai2017scannet}
Angela Dai, Angel~X Chang, Manolis Savva, Maciej Halber, Thomas Funkhouser, and Matthias Nie{\ss}ner.
\newblock Scannet: Richly-annotated 3d reconstructions of indoor scenes.
\newblock In {\em Proceedings of the IEEE conference on computer vision and pattern recognition}, pages 5828--5839, 2017.

\bibitem{glrv}
Hehe Fan, Xiaojun Chang, Wanyue Zhang, Yi Cheng, Ying Sun, and Mohan Kankanhalli.
\newblock Self-supervised global-local structure modeling for point cloud domain adaptation with reliable voted pseudo labels.
\newblock In {\em Proceedings of the IEEE/CVF Conference on Computer Vision and Pattern Recognition}, pages 6377--6386, 2022.

\bibitem{dann}
Yaroslav Ganin, Evgeniya Ustinova, Hana Ajakan, Pascal Germain, Hugo Larochelle, Fran{\c{c}}ois Laviolette, Mario Marchand, and Victor Lempitsky.
\newblock Domain-adversarial training of neural networks.
\newblock {\em The journal of machine learning research}, 17(1):2096--2030, 2016.

\bibitem{unified_adapter}
Junxian He, Chunting Zhou, Xuezhe Ma, Taylor Berg-Kirkpatrick, and Graham Neubig.
\newblock Towards a unified view of parameter-efficient transfer learning.
\newblock In {\em International Conference on Learning Representations}, 2022.

\bibitem{cycada}
Judy Hoffman, Eric Tzeng, Taesung Park, Jun-Yan Zhu, Phillip Isola, Kate Saenko, Alexei Efros, and Trevor Darrell.
\newblock Cycada: Cycle-consistent adversarial domain adaptation.
\newblock In {\em International conference on machine learning}, pages 1989--1998. Pmlr, 2018.

\bibitem{adapter}
Neil Houlsby, Andrei Giurgiu, Stanislaw Jastrzebski, Bruna Morrone, Quentin De~Laroussilhe, Andrea Gesmundo, Mona Attariyan, and Sylvain Gelly.
\newblock Parameter-efficient transfer learning for nlp.
\newblock In {\em International Conference on Machine Learning}, pages 2790--2799. PMLR, 2019.

\bibitem{kang2019contrastive}
Guoliang Kang, Lu Jiang, Yi Yang, and Alexander~G Hauptmann.
\newblock Contrastive adaptation network for unsupervised domain adaptation.
\newblock In {\em Proceedings of the IEEE/CVF Conference on Computer Vision and Pattern Recognition}, pages 4893--4902, 2019.

\bibitem{gcn}
Thomas~N. Kipf and Max Welling.
\newblock Semi-supervised classification with graph convolutional networks.
\newblock In {\em International Conference on Learning Representations}, 2017.

\bibitem{mlsp}
Hanxue Liang, Hehe Fan, Zhiwen Fan, Yi Wang, Tianlong Chen, Yu Cheng, and Zhangyang Wang.
\newblock Point cloud domain adaptation via masked local 3d structure prediction.
\newblock In {\em European Conference on Computer Vision}, pages 156--172. Springer, 2022.

\bibitem{rscnn}
Yongcheng Liu, Bin Fan, Shiming Xiang, and Chunhong Pan.
\newblock Relation-shape convolutional neural network for point cloud analysis.
\newblock In {\em Proceedings of the IEEE/CVF conference on computer vision and pattern recognition}, pages 8895--8904, 2019.

\bibitem{mom}
Karl Pearson.
\newblock Method of moments and method of maximum likelihood.
\newblock {\em Biometrika}, 28(1/2):34--59, 1936.

\bibitem{pointnet}
Charles~R Qi, Hao Su, Kaichun Mo, and Leonidas~J Guibas.
\newblock Pointnet: Deep learning on point sets for 3d classification and segmentation.
\newblock In {\em Proceedings of the IEEE conference on computer vision and pattern recognition}, pages 652--660, 2017.

\bibitem{pointnet++}
Charles~Ruizhongtai Qi, Li Yi, Hao Su, and Leonidas~J Guibas.
\newblock Pointnet++: Deep hierarchical feature learning on point sets in a metric space.
\newblock {\em Advances in neural information processing systems}, 30, 2017.

\bibitem{pointnext}
Guocheng Qian, Yuchen Li, Houwen Peng, Jinjie Mai, Hasan Abed Al~Kader Hammoud, Mohamed Elhoseiny, and Bernard Ghanem.
\newblock Pointnext: Revisiting pointnet++ with improved training and scaling strategies.
\newblock {\em arXiv preprint arXiv:2206.04670}, 2022.

\bibitem{pointdan}
Can Qin, Haoxuan You, Lichen Wang, C-C~Jay Kuo, and Yun Fu.
\newblock Pointdan: A multi-scale 3d domain adaption network for point cloud representation.
\newblock {\em Advances in Neural Information Processing Systems}, 32, 2019.

\bibitem{rs}
Jonathan Sauder and Bjarne Sievers.
\newblock Self-supervised deep learning on point clouds by reconstructing space.
\newblock {\em Advances in Neural Information Processing Systems}, 32, 2019.

\bibitem{shen2022domain}
Yuefan Shen, Yanchao Yang, Mi Yan, He Wang, Youyi Zheng, and Leonidas~J Guibas.
\newblock Domain adaptation on point clouds via geometry-aware implicits.
\newblock In {\em Proceedings of the IEEE/CVF Conference on Computer Vision and Pattern Recognition}, pages 7223--7232, 2022.

\bibitem{vl_adapter}
Yi-Lin Sung, Jaemin Cho, and Mohit Bansal.
\newblock Vl-adapter: Parameter-efficient transfer learning for vision-and-language tasks.
\newblock In {\em Proceedings of the IEEE/CVF Conference on Computer Vision and Pattern Recognition (CVPR)}, pages 5227--5237, June 2022.

\bibitem{kpconv}
Hugues Thomas, Charles~R Qi, Jean-Emmanuel Deschaud, Beatriz Marcotegui, Fran{\c{c}}ois Goulette, and Leonidas~J Guibas.
\newblock Kpconv: Flexible and deformable convolution for point clouds.
\newblock In {\em Proceedings of the IEEE/CVF international conference on computer vision}, pages 6411--6420, 2019.

\bibitem{adapt_segmap}
Yi-Hsuan Tsai, Wei-Chih Hung, Samuel Schulter, Kihyuk Sohn, Ming-Hsuan Yang, and Manmohan Chandraker.
\newblock Learning to adapt structured output space for semantic segmentation.
\newblock In {\em Proceedings of the IEEE conference on computer vision and pattern recognition}, pages 7472--7481, 2018.

\bibitem{adda}
Eric Tzeng, Judy Hoffman, Kate Saenko, and Trevor Darrell.
\newblock Adversarial discriminative domain adaptation.
\newblock In {\em Proceedings of the IEEE conference on computer vision and pattern recognition}, pages 7167--7176, 2017.

\bibitem{vaswani2017attention}
Ashish Vaswani, Noam Shazeer, Niki Parmar, Jakob Uszkoreit, Llion Jones, Aidan~N Gomez, {\L}ukasz Kaiser, and Illia Polosukhin.
\newblock Attention is all you need.
\newblock {\em Advances in neural information processing systems}, 30, 2017.

\bibitem{dgcnn}
Yue Wang, Yongbin Sun, Ziwei Liu, Sanjay~E Sarma, Michael~M Bronstein, and Justin~M Solomon.
\newblock Dynamic graph cnn for learning on point clouds.
\newblock {\em Acm Transactions On Graphics (tog)}, 38(5):1--12, 2019.

\bibitem{wu20153d}
Zhirong Wu, Shuran Song, Aditya Khosla, Fisher Yu, Linguang Zhang, Xiaoou Tang, and Jianxiong Xiao.
\newblock 3d shapenets: A deep representation for volumetric shapes.
\newblock In {\em Proceedings of the IEEE conference on computer vision and pattern recognition}, pages 1912--1920, 2015.

\bibitem{pointtransformer}
Hengshuang Zhao, Li Jiang, Jiaya Jia, Philip~HS Torr, and Vladlen Koltun.
\newblock Point transformer.
\newblock In {\em Proceedings of the IEEE/CVF International Conference on Computer Vision}, pages 16259--16268, 2021.

\bibitem{gast}
Longkun Zou, Hui Tang, Ke Chen, and Kui Jia.
\newblock Geometry-aware self-training for unsupervised domain adaptation on object point clouds.
\newblock In {\em Proceedings of the IEEE/CVF International Conference on Computer Vision}, pages 6403--6412, 2021.

\end{thebibliography}
}


\end{document}


\title{Supplementary Material for \\ PC-Adapter: Topology-Aware Adapter for Efficient Domain Adaption on Point Clouds  with Rectified Pseudo-label}
\maketitle

\ificcvfinal\thispagestyle{empty}\fi
\appendix
\section{Additional Experimental Results}

\subsection{Further Discovery of Remark 3.2}
We provide additional examples that Point Transformer~\cite{pointtransformer} misclassifies target domain objects as ground-truth classes of source domain objects in Figure~\ref{fig:pointtransformer_misprediction_appendix}. The figure illustrates that the existing encoding architecture fails to perceive the implicit shape of target objects in its entirety, as described in Remark \textcolor{red}{3.2}. Specifically, the encoder exhibits poor prediction ability on target objects due to its \textit{partial} focus on similar outlines (\ie cylinders in lamps and stems in plants) as shown in Figure~\ref{fig:pointtransformer_misprediction_appendix} (a) and the presence of multiple legs \hjs{in} Figure~\ref{fig:pointtransformer_misprediction_appendix} (b).
\begin{figure}[h] 
\centering
   \includegraphics[width=0.488\textwidth]{figures/appendix_A_1_cropped.pdf}
  \vspace{-0.1in}
  \caption{\small{\textcolor{blue}{\textbf{Target}} point cloud samples that Point Transformer mispredicts as the ground-truth classes of \textcolor{teal}{\textbf{source}} samples on two domain shift scenarios - (a) ShapeNet$\rightarrow$ModelNet and (b) ScanNet$\rightarrow$ShapeNet. }}
  \label{fig:pointtransformer_misprediction_appendix}
  \vspace{-0.1in}
\end{figure}
\subsection{Learning Rate Adjustment for Different Training Paths}~\label{subsec:appx_gradual_LR}
As we describe in Section \textcolor{red}{3.2} of our main paper, we reduce the learning rate by a factor of $\rho$ for the parameters of the shared components ($\Phi$ and $\Psi_g$) to preserve the implicit shape knowledge of the source domain while training on target point clouds. To demonstrate the effectiveness of this strategy, we compare domain adaptation results of ours with and \textit{without} weakly updating parameters of shared parts for the target path (Table~\ref{tb:gradual_lr}). The results show that maintaining an identical learning rate for both the source and target training paths significantly degrades the adaptation performance. 
These results highlight the importance of altering the learning rate on source and target paths, which is effective for \textit{preserving }the source geometry information.

\begin{table}[h]
\caption{\small Ablation study for learning rate adjustment strategy on two domain shift settings in PointDA-10, averaged over three repetitions ($\pm$ SEM).} 
\vspace{-0.1in}
\begin{center}
\begin{footnotesize}
\setlength{\columnsep}{1pt}%
\begin{adjustbox}{width=0.8\linewidth}
\begin{tabular}{@{\extracolsep{1pt}}ccc}
\toprule
\multirow{2}{*}{\textbf{Method}} & \multirow{1}{*}{S$\rightarrow$M} & \multirow{1}{*}{M$\rightarrow$S*}  \\
\cline{2-3}
 & Acc. & Acc. \\ 
\cline{1-3}
                \textit{PC-Adapter} ($\rho=1.0$)  & 74.8 \tiny{$\pm 0.4$} & 53.6 \tiny{$\pm 0.5$} \\
                \textit{PC-Adapter} ($\rho=\frac{1}{2.5}$)  & 76.6 \tiny{$\pm 0.3$} & 53.9 \tiny{$\pm 0.2$} \\
                \textit{\textbf{PC-Adapter}} ($\rho=\frac{1}{5.0}$) & \textbf{77.5} \tiny{$\pm 0.2$} & \textbf{58.2} \tiny{$\pm 0.4$} \\
\bottomrule
\end{tabular}
\end{adjustbox}
\end{footnotesize}
\end{center}
\label{tb:gradual_lr}
\vspace{-0.2in}
\end{table}

\subsection{Qualitative Analysis of Relative Positional Encoding} 
In Figure~\ref{fig:remark2_figure_appendix}, we qualitatively analyze target samples to complement the efficacy of the proposed \textit{relative} positional encoding $\sigma$. The target objects in the figure are ones that \textit{PC-Adapter} equipped with conventional point cloud positional encoding~\cite{pointtransformer} mispredicts whereas \textit{PC-Adapter} with relative positional encoding correctly classifies. 

\section{Proof: Parameter Estimation for Beta Distribution}
Before proving the parameter estimation for beta distribution, we first introduce the method-of-moment\hjs{s} estimation below.

\noindent\textbf{Lemma 1} (Method of Moments). \quad
Let $x=\{x_1, \dots, x_n\}$ be a set of independent and identically distributed realizations (samples) from random variable $\bm{X}$. We define the probability distribution $p(x|\theta)$ parameterized by unknown parameters $\theta = 
\{\theta_1, \dots, \theta_k\}$. Then the unknown parameters are estimated by matching the moments as follows. \\

\noindent Let us derive first $k$ sample moments $\{\hat{\mu}_i(\bm{X})\}_{i=1}^{k}$ for constant $b_k$ as
\begin{equation}
\hat{\mu}_1(\bm{X}) =\frac{1}{n}\sum_{i=1}^n(x_i - b_1)^1,\ \dots \ , \hat{\mu}_k(\bm{X}) =\frac{1}{n}\sum_{i=1}^n(x_i - b_k)^k.
\end{equation}
\noindent Then we express the first $k$ moments of $\bm{X}$ in terms of $\theta$:
\begin{equation}
{\mu}_1(\bm{X}) = f_1(\theta_1, \dots, \theta_k ),\ \dots \ , {\mu}_k(\bm{X}) = f_k(\theta_1, \dots, \theta_k ).
\end{equation}

\noindent By solving the following system of $k$ equations,
\begin{equation}
\begin{cases}
\hat{\mu}_1(\bm{X})=f_1(\theta_1, \dots, \theta_k ) \\
\vdots \\
\hat{\mu}_k(\bm{X})=f_k(\theta_1, \dots, \theta_k ),
\end{cases}
\end{equation}

\noindent the estimated parameters $\hat{\theta}_1, \dots, \hat{\theta}_k$ can be derived.

Assume the confidence distribution for each class $t$ given source training dataset $\mathcal{S}_\text{train}$ follows a beta distribution, $p(c_t|\mathcal{S}_{\text{train}}) \approx \texttt{Beta}(\hat{\alpha}_t, \hat{\beta}_t)$. 
For each class $t$,  we compute the sample mean of confidences $\bar{c}_t$ as $\bar{c}_t = \frac{1}{|\mathcal{S}^{t}_{\text{train}}|}\sum_{i\in \mathcal{S}^{t}_{\text{train}}}c_{i,t}$, and sample variance $\bar{v}_t$ as $\bar{v}_t = \frac{1}{|\mathcal{S}^{t}_{\text{train}}|-1}\sum_{i\in \mathcal{S}^{t}_{\text{train}}}(c_{i,t}-\bar{c}_t)^2$, where  $\mathcal{S}^t_{\text{train}}$ denotes the indices of samples belonging to class $t$, and $c_{i,t}$ is the confidence score for class $t$ on the $i$-th sample. Then, two unknown parameters - $\hat{\alpha}_t$ and $\hat{\beta}_t$ - are estimated as:
\begin{equation}
\hat{\alpha}_{t} = \bar{c}_t\big(\frac{\bar{c}_t(1-\bar{c}_t)}{\bar{v}_t}-1\big), \ \hat{\beta}_{t} = (1-\bar{c}_t)\big(\frac{\bar{c}_t(1-\bar{c}_t)}{\bar{v}_t}-1\big).  
\end{equation}
 \noindent\textit{Proof.}
Mean and variance of random variable $\bm{X}$ which follows beta distribution $\texttt{Beta}(\hat{\alpha}_t, \hat{\beta}_t)$ are given by
\begin{equation}
\mathrm{E}(\bm{X}) = \frac{\hat{\alpha}_t}{\hat{\alpha}_t+\hat{\beta}_t}, \mathrm{Var}(\bm{X})= \frac{\hat{\alpha}_t\hat{\beta}_t}{(\hat{\alpha}_t+\hat{\beta}_t)^2(\hat{\alpha}_t+\hat{\beta}_t+1)}.
\end{equation}

\noindent Using Lemma 1, we acquire the following equations by matching the moments:
\begin{gather}
 \bar{c}_t=\frac{\hat{\alpha}_t}{\hat{\alpha}_t+\hat{\beta}_t}~\label{eq6} \\
 \bar{v}_t=\frac{\hat{\alpha}_t\hat{\beta}_t}{(\hat{\alpha}_t+\hat{\beta}_t)^2(\hat{\alpha}_t+\hat{\beta}_t+1)}~\label{eq7} 
\end{gather}

\noindent From the Equation~\ref{eq6},
\begin{align}
 \bar{c}_t(\hat{\alpha}_t+\hat{\beta}_t) &= \hat{\alpha}_t \nonumber \\
 \bar{c}_t\hat{\beta}_t &= \hat{\alpha}_t -\bar{c}_t\hat{\alpha}_t \nonumber \\
\hat{\beta}_t &= \hat{\alpha}_t(\frac{1}{\bar{c}_t}-1). ~\label{eq8}
\end{align}
Plugging Equation~\ref{eq8} into Equation~\ref{eq7}, we have:
\begin{align}
 \bar{v}_t&=\frac{\hat{\alpha}_{t}^{2}(\frac{1}{\bar{c}_t}-1)}{(\frac{\hat{\alpha}_{t}}{\bar{c}_t})^{2}(\frac{\hat{\alpha}_{t}}{\bar{c}_t}+1) } \nonumber\\
 &= \frac{(\frac{1}{\bar{c}_t}-1)}{(\frac{1}{\bar{c}_t})^{2}(\frac{\hat{\alpha}_{t}}{\bar{c}_t}+1)} \nonumber \\
 &= \frac{\bar{c}_t-\bar{c}_{t}^{2}}{\frac{\hat{\alpha}_{t}}{\bar{c}_t}+1} \nonumber \\
 \bar{v}_t(\frac{\hat{\alpha}_{t}}{\bar{c}_t}+1) &= \bar{c}_t-\bar{c}_{t}^{2} \nonumber \\
 \frac{\hat{\alpha}_{t}}{\bar{c}_t}+1 &= \frac{\bar{c}_t-\bar{c}_{t}^{2}}{\bar{v}_t} \nonumber \\
 \hat{\alpha}_{t} &= \bar{c}_t\big(\frac{\bar{c}_t(1-\bar{c}_{t})}{\bar{v}_t} -1\big).~\label{eq9} 
\end{align}
We can also obtain $\hat{\beta}_{t}$ using Equation \ref{eq8} and Equation~\ref{eq9}:
\begin{align}
\hat{\beta}_t &= \hat{\alpha}_t(\frac{1}{\bar{c}_t}-1) \nonumber \\
&= \bar{c}_t\big(\frac{\bar{c}_t(1-\bar{c}_{t})}{\bar{v}_t} -1\big)(\frac{1}{\bar{c}_t}-1) \nonumber\\
&= (1-\bar{c}_t)\big(\frac{\bar{c}_t(1-\bar{c}_t)}{\bar{v}_t}-1\big),
\end{align}
which ends the proof. \hfill$\qedsymbol$

\section{Training Algorithm of \textit{PC-Adapter}}
We provide detailed algorithm of \textit{PC-Adapter} in Algorithm~\ref{Alg1:training_procedure}.
\begin{algorithm*}[t]
    \caption{Overall training procedure of \textit{PC-Adapter}.}
    \setstretch{1.05}
    \begin{algorithmic}[1]
        \Require{Source data $\mathcal{S}=\{(\bm{X}_{k}^{\text{src}}=\{\mathbf{p}_i\}_{i=1}^m,y_{k}^{\text{src}})\}_{k=1}^{n_s}$, target data $\mathcal{T}=\{(\bm{X}_{k}^{\text{trgt}}\}=\{\mathbf{p}_i'\}_{i=1}^m)\}_{k=1}^{n_t}$, feature encoder $\Phi$, shape-aware adapter $\Psi_g$, locality-aware adapter $\Psi_l$, classifier $f$, correction intensity $r_0$}.
        
        \For{$e = 1 \ldots E$}
            \For{$(\{\mathbf{p}_i\}_{i=1}^m,y^{\text{src}}_k),\{\mathbf{p}_i'\}_{i=1}^m$ in $(\mathcal{S,T})$}
            \State Obtain source encoder output $\{\Phi(\mathbf{p}_i)\}_{i=1}^m$
            \State Sample farthest points $\bm{\tilde X}_{k}^{\text{src}}=\{\mathbf{p}_i\}_{i=1}^{m'}$
            \State $\{\Psi_g(\mathbf{p}_i)\}_{i=1}^{m'}\leftarrow \sum_{\mathbf{p}_j\in\bm{\tilde X}_k^{\text{src}}\backslash\mathbf p_i} w_{ij}(\varphi(\Phi(\mathbf{p}_j))+\sigma_{ij})$ \Comment{Encoding by shape-aware adapter}
            \State Train $f$ on $\texttt{Combine} \big(\{\Psi_g(\mathbf{p}_i)\}_{i=1}^{m'},\{\Phi(\mathbf{p}_i)\}_{i=1}^{m}\big)$ with label $y^{\text{src}}_k$ \\
            \For{$t = 1 \ldots c$} 
                \State $\bar{c}_t \leftarrow \frac{1}{|\mathcal{S}^{t}_{\text{train}}|}\sum_{i\in \mathcal{S}^{t}_{\text{train}}}c_{i,t}$,  $ \ \bar{v}_t \leftarrow \frac{1}{|\mathcal{S}^{t}_{\text{train}}|-1}\sum_{i\in \mathcal{S}^{t}_{\text{train}}}(c_{i,t}-\bar{c}_t)^2 $
                \State $\hat{\alpha}_{t} \leftarrow \bar{c}_t\big(\frac{\bar{c}_t(1-\bar{c}_t)}{\bar{v}_t}-1\big)$
                \State $\hat{\beta}_{t} \leftarrow(1-\bar{c}_t)\big(\frac{\bar{c}_t(1-\bar{c}_t)}{\bar{v}_t}-1\big)$
                \State Compute $r_i$ from percentile point function of $\texttt{Beta}(\hat{\alpha}_t, \hat{\beta}_t)$
                \State $\tilde{c}_{i,t}\leftarrow c_{i,t}\cdot\big(\frac{1}{1-r_i+r_0}\big)$ \Comment{Adjusting confidence score for each class}
            \EndFor
            \State $\hat{y}_k^{\text{trgt}}\leftarrow \underset{t}{\operatorname{argmax}}\;\tilde{c}_{i,t}$ \\
            \State Repeat line 3 - 5 on $\{(\mathbf{p}_i')\}_{i=1}^m$
            \State $\{\Psi_l(\mathbf{p}_i')\}_{i=1}^{m'}\leftarrow \mathbf{\Theta}^{\mathsf{T}}\sum_{\scriptstyle{\mathbf{p}_j\in \mathcal{N}(i)\cup\{\mathbf{p}_i\}}}\frac{e_{j,i}}{\texttt{deg}(\mathbf{p}_j)\texttt{deg}(\mathbf{p}_i)}\Phi(\mathbf{p}_j)$ \Comment{Encoding by locality-aware adapter}
            \State Train $f$ on $\texttt{Combine} \big(\{\Psi_g(\mathbf{p}_i)\}_{i=1}^{m'}, \{\Psi_l(\mathbf{p}_i)\}_{i=1}^{m'}, \{\Phi(\mathbf{p}_i)\}_{i=1}^{m'}\big)$ with label $\hat{y}_k^{\text{trgt}}$
            \EndFor
        \EndFor
\end{algorithmic}
\label{Alg1:training_procedure}
\end{algorithm*}

\begin{table*}[t] 
\center

\caption{\small Data statistics. The number of samples for each class in PointDA-10.}
\vspace{-0 in}
\begin{small}
\setlength{\tabcolsep}{3pt} 
\begin{tabular}{lcc|ccccccccccc}
\toprule
    \textbf{Dataset} & \textbf{Domain}  & & Bathtub & Bed & Bookshelf & Cabinet & Chair & Lamp & Monitor & Plant & Sofa & Table & \textbf{Total}  \\
    \hline
    \multirow{2}{*}{ModelNet-10} & \multirow{2}{*}{Synthetic} & Train & 106 & 515 & 572 & 200 & 889 & 124 & 465 & 240 & 680 & 392 & 4,183  \\ 
     & & Test & 50 & 100 & 100 & 86 & 100 & 20 & 100 & 100 & 100 & 100  & 856 \\
    \cline{1-14}
    \multirow{2}{*}{ShapeNet-10} & \multirow{2}{*}{Synthetic} & Train & 599 & 167 & 310 & 1,076 & 4,612 & 1,620 & 762 & 158 & 2,198 & 5,876 & 17,378 \\
    & & Test & 85 & 23 & 50 & 126 & 662 & 232 & 112 & 30 & 330 & 842 & 2,492   \\
    \cline{1-14}
    \multirow{2}{*}{ScanNet-10} & \multirow{2}{*}{Real} & Train & 98 & 329 & 464 & 650 & 2,578 & 161 & 210 & 88 & 495 & 1,037 & 6,110  \\
    & & Test & 26 & 85 & 146 & 149 & 801 & 41 & 61 & 25 & 134 & 301 & 1,769 \\
\bottomrule
\end{tabular}
\end{small}
\label{tb:datastat_pointda}

\end{table*}
\section{Detailed Experiment Setup}
\subsection{Dataset Statistics}
\paragraph{PointDA-10}
The PointDA-10~\cite{pointdan} dataset comprises objects from 10 shared classes that are collected in three datasets - ModelNet40 (\textbf{M})~\cite{wu20153d}, ShapeNet (\textbf{S}) ~\cite{chang2015shapenet}, and ScanNet (\textbf{S*})~\cite{dai2017scannet}. As provided in Table~\ref{tb:datastat_pointda}, the ModelNet-10 contains 4,183 training and 856 test point clouds, while ShapeNet-10 consists of 17,378 training and 2,492 test samples. Point clouds of both datasets are acquired by sampling 2,048 points from the surface of 3D CAD models (\ie synthetic datasets). Compared to these datasets, ScanNet-10 includes 6,110 training and 1,769 point clouds with 2048 points each, which are scanned and reconstructed from real-world scenes. The point clouds in ScanNet-10 usually exhibit partial views of objects due to the occlusion (by adjacent objects or self-occlusion) and sensor noises.

We carefully follow the data preparation and data split procedures adopted in previous studies~\cite{pointdan,defrec,gast,shen2022domain}. The object point clouds from all datasets are oriented to align with the direction of gravity, while arbitrary rotations along the $z$-axis are tolerated. Then, point clouds are normalized to fit within a unit cube. Input point clouds with 2,048 points are down-sampled to 1,024 points.

\paragraph{GraspNetPC-10}
The point clouds in GraspNetPC-10~\cite{shen2022domain,fang2020graspnet} are collected through raw depth scans on both real-world and synthetic scenes using two different cameras, Kinect2 and Intel Realsense. Specifically, the raw depth scans are projected to 3D space and object segmentation masks are applied to extract the object point clouds from the scenes. Unlike samples in PointDA-10, point clouds of this dataset are not aligned along with the vertical direction. For real scenes, the Kinect domain includes 10,973 training and 2,560 test samples, while the Realsense domain contains 10,698 training and 2,560 test point clouds. 
For synthetic scenes, the Kinect domain has 10,887 training samples, and the Realsense domain contains 10,542 training point clouds. Detailed statistics are provided in Table~\ref{tb:datastat_graspnet}. Point clouds captured by the two different cameras are often affected by different types of noises, and different degrees of structural distortions. We follow the data preparation procedure of \cite{shen2022domain}.

\begin{table}[h] 
\center
\vspace{-0.02in}
\caption{\small  Data statistics. The number of samples for each class in GraspNetPC-10. Syn. and Real denote synthetic scenes and real scenes, respectively.}
\vspace{-0.00in}
\begin{footnotesize}
\setlength{\tabcolsep}{.8pt} 
\begin{adjustbox}{width=0.98\linewidth}
\begin{tabular}{lc|ccccccccccc}
\toprule
     \textbf{Domain}&  & $\mathbf{L}_0$ & $\mathbf{L}_1$ & $\mathbf{L}_2$ & $\mathbf{L}_3$ & $\mathbf{L}_4$ & $\mathbf{L}_5$ & $\mathbf{L}_6$ & $\mathbf{L}_7$ & $\mathbf{L}_8$ & $\mathbf{L}_9$ & \textbf{Total } \\
    \hline
    Kinect  & \multirow{2}{*}{Train} & \multirow{2}{*}{1,024} & \multirow{2}{*}{1,280} & \multirow{2}{*}{1,273} & \multirow{2}{*}{1,024} & \multirow{2}{*}{1,020} & \multirow{2}{*}{1,278} & \multirow{2}{*}{934} & \multirow{2}{*}{1,006} & \multirow{2}{*}{ 1,024} & \multirow{2}{*}{1,024} & \multirow{2}{*}{10,887} \\
    (Syn.) & &&&&&&&&&&& \\
    \cline{1-13}
    Realsense  & \multirow{2}{*}{Train} & \multirow{2}{*}{972} & \multirow{2}{*}{1,280} & \multirow{2}{*}{1,280} &  \multirow{2}{*}{1,024} &  \multirow{2}{*}{1,024} & \multirow{2}{*}{1,280} & \multirow{2}{*}{895} & \multirow{2}{*}{792} & \multirow{2}{*}{980} & \multirow{2}{*}{1,015} & \multirow{2}{*}{10,542} \\
    (Syn.) & &&&&&&&&&&& \\
    \cline{1-13}
    Kinect & Train &  1,024 & 1,280 & 1,273 &  1,024 & 1,015 & 1,272 & 1,019 &  1,024 & 1,018 &  1,024 & 10,973 \\
    (Real)& Test & 256 & 256 & 256 & 256 & 256 & 256 & 256 & 256 & 256 & 256 & 2,560 \\
    \cline{1-13}
    Realsense & Train & 968 & 1,280 & 1,280 &  1,024 & 1,020 & 1,279 & 1,020 & 841 & 971 & 1,015 & 10,698 \\
    (Real) & Test & 256 & 256 & 256 & 256 & 256 & 256 & 256 & 256 & 256 & 256 & 2,560 \\
\bottomrule
\end{tabular}
\end{adjustbox}
\end{footnotesize}
\label{tb:datastat_graspnet}
\vspace{-0.05in}
\end{table}

\paragraph{PointSegDA}
The PointSegDA dataset~\cite{defrec} originates from a 3D mesh-structured human model dataset~\cite{mesh_segda}, featuring four distinct subsets. These subsets encompass a total of eight human body part classes, exhibiting variations in terms of point distribution, pose, and scanned individuals. The process involves generating point cloud data by sampling 2048 points from the 3D mesh data. Subsequently, the sampled points are aligned along the $z$-axis and normalized to fit within a unit cube. Corresponding point labels are assigned based on polygon labels. Concise data statistics are presented in Table~\ref{tb:datastat_pointsegda}. We adhere to the data preprocessing and data split rules proposed in \cite{defrec}.

\begin{table}[h] 
\center
\vspace{-0.02in}
\caption{\small  Data statistics. The number of samples for each subset in PointSegDA.}
\vspace{-0.00in}
\begin{footnotesize}
\setlength{\tabcolsep}{1.2pt} 
\begin{adjustbox}{width=0.65\linewidth}
\begin{tabular}{l|cccc}
\toprule
     \textbf{Domain} & Train & Validation & Test &  \textbf{Total} \\
     \hline
     FAUST & 70 & 10 & 20 & 100 \\
     MIT & 118 & 17 & 34 & 169 \\
     ADOBE & 29 & 4 & 8 & 41 \\
     SCAPE & 50 & 7 & 14 & 71\\

\bottomrule
\end{tabular}
\end{adjustbox}
\end{footnotesize}
\label{tb:datastat_pointsegda}
\vspace{-0.05in}
\end{table}

\subsection{Implementation Details}
In this subsection, we describe the implementation details and hyperparameters of our method. For our experiments, we set the farthest point sampling (FPS) ratio in to 0.1 for adapter modules, and the number of nearest neighbors $k$ to 5 for $k$-NN graph in locality-aware adapter $\Psi_l$. For our distribution-guided pseudo labeling, we tune the correction intensity, $r_0$, from the $\{0.1, 10, 15, 20, 30, 40, 45\}$. As in previous works~\cite{glrv,gast}, we employ the (fixed) threshold, $\gamma$, to filter out noisy pseudo labels that have low confidence scores. We search optimal $\gamma$ among the range $[0.7,0.92]$. Since our correction strategy modifies the confidences by a ratio from $\frac{1}{r_0+1}$ to $\frac{1}{r_0}$, we multiply the average scale of modification to threshold $\gamma$, $\gamma * \frac{1}{2}(\frac{1}{r_0+1}+\frac{1}{r_0})$, to account for the altered scale of confidence scores. The loss coefficient for the regularization loss $\mathcal{L}_{\text{centroid}}$ is tuned from the set \{1, 0.1, 0.001\}. In part segmentation experiments, we do not use distribution-guided pseudo labels and select a threshold $\gamma$ from the options \{0.98, 0.99\}.

\subsection{Evaluation Protocol}
In our experiments, we use DGCNN~\cite{dgcnn} as the architecture for the feature encoder $\Phi$, in line with previous works~\cite{glrv,shen2022domain}. We follow the evaluation protocol outlined in \cite{glrv,defrec} for the PointDA-10 dataset and that in~\cite{shen2022domain} for the GraspNetPC-10 dataset. For point cloud classification experiments, we adopt an Adam optimizer with an initial learning rate 0.001, and weight decay is set to 0.00005. The cosine annealing is employed for the learning rate scheduler and the learning rate weakening factor, $\rho$, is set to 0.2 for target domain training, except for the ShapeNet to ScanNet setting, where it is set to 0.01. We train the models for 150 epochs on PointDA-10 and 120 epochs on GraspNetPC-10, and the best model is selected using source validation accuracy. For the PointSegDA dataset, we follow the evaluation settings of \cite{defrec}. PCM data augmentation is employed in scenarios where it is utilized in \cite{defrec}. In part segmentation experiments, we search the learning rate decay factor $\rho$ within the set \{$1$, $\frac{1}{3}$, $\frac{1}{5}$\}.

\begin{figure*}[h] 
\centering
    \includegraphics[width=0.6\textwidth]{figures/appendix_remark2_figure_cropped_cameraready.pdf}
  \vspace{-0.0in}
  \caption{{\textcolor{blue}{\textbf{Target}} point cloud samples that proposed relative positional encoding \textit{correctly} predicts while traditional point cloud positional encoding misclassifies as the ground-truth classes of \textcolor{teal}{\textbf{source}} samples on two domain shift scenarios - (a) target \textit{beds} mispredicted as bathtub, chair, and sofa, (b) target \textit{tables} mispredicted as bed and sofa of the source. The results indicate that the major drawbacks of existing encoding architectures discussed in Section \textcolor{red}{3.2} and \textcolor{red}{A.1} could be effectively mitigated by our proposed \textit{relative} positional encoding.}}
  \label{fig:remark2_figure_appendix}

\end{figure*}

{\small
\bibliographystyle{ieee_fullname}
\bibliography{iccv2023}
}